\ifdefined\pdfminorversion\pdfminorversion=7\fi
\documentclass[10pt,letterpaper]{slai_preprint}
\usepackage[numbers,square,sort]{natbib}
\setcitestyle{numbers,square,citesep={,}}
\usepackage{amsmath,amssymb}
\usepackage{algorithm,algpseudocode}
\usepackage{graphicx}
\usepackage{booktabs}
\usepackage{xcolor}
\usepackage{array,tabularx}
\usepackage{multirow}
\usepackage{placeins}

\newcommand{\tablehead}[1]{\textbf{#1}}
\newcommand{\method}{EVO-WAM}
\newcommand{\methodbackbone}[1]{\mbox{\method{}\textsubscript{#1}}}
\usepackage{hyperref}
\hypersetup{colorlinks=true,linkcolor=slaiHeading,citecolor=slaiHeading,
  urlcolor=slaiHeading,breaklinks=true,
  pdftitle={EVO-WAM: Evolving World Action Models through Video-Action Verification},
  pdfauthor={Shiyang Zhou, Xionghao Wu, Wenbo Li, Shenghe Zheng, Jiyao Zhang, Songsong Yu, Yijun Yang, Jianhui Liu, Haoze Sun, Senqiao Yang, Li Jiang, Jingyong Su, Haoyang Huang, Zhuotao Tian}}
\usepackage{url}

\title{EVO-WAM: Evolving World Action Models\\through Video-Action Verification}
\author{%
  \mbox{Shiyang Zhou\textsuperscript{1,2,\ensuremath{\ddagger}}},
  \mbox{Xionghao Wu\textsuperscript{3,\ensuremath{\ddagger}}},
  \mbox{Wenbo Li\textsuperscript{4,\ensuremath{\dagger}}},
  \mbox{Shenghe Zheng\textsuperscript{5}},
  \mbox{Jiyao Zhang\textsuperscript{6}},
  \mbox{Songsong Yu\textsuperscript{7}},
  \mbox{Yijun Yang\textsuperscript{5,8}},
  \mbox{Jianhui Liu\textsuperscript{9}},
  \mbox{Haoze Sun\textsuperscript{10}},
  \mbox{Senqiao Yang\textsuperscript{11}},\\
  \mbox{Li Jiang\textsuperscript{12,2}},
  \mbox{Jingyong Su\textsuperscript{1}},
  \mbox{Haoyang Huang\textsuperscript{4}},
  \mbox{Zhuotao Tian\textsuperscript{1,2,*}}}
\affiliations{%
  \textsuperscript{1}HITSZ\quad
  \textsuperscript{2}SLAI\quad
  \textsuperscript{3}THU\quad
  \textsuperscript{4}JD\quad
  \textsuperscript{5}HKUST\quad
  \textsuperscript{6}PKU\\
  \textsuperscript{7}SJTU\quad
  \textsuperscript{8}HKUSTGZ\quad
  \textsuperscript{9}HKU\quad
  \textsuperscript{10}UBC\quad
  \textsuperscript{11}CUHK\quad
  \textsuperscript{12}CUHKSZ}
\authoremails{%
  \href{mailto:shiyangzhou@stu.hit.edu.cn}{\texttt{shiyangzhou@stu.hit.edu.cn}}\quad
  \href{mailto:fenglinglwb@gmail.com}{\texttt{fenglinglwb@gmail.com}}\quad
  \href{mailto:tianzhuotao@hit.edu.cn}{\texttt{tianzhuotao@hit.edu.cn}}}
\authornote{\textsuperscript{\ensuremath{\ddagger}}Equal contribution
\quad
  \textsuperscript{\ensuremath{\dagger}}Project lead\quad
  \textsuperscript{*}Corresponding author}
\newcommand{\BaseAverage}{31.6}
\newcommand{\OursAverage}{68.0}
\newcommand{\StartAverage}{26.9}
\newcommand{\SeenBase}{85.8}
\newcommand{\SeenFinal}{84.8}

\newcommand{\OursSceneB}{70.4}
\newcommand{\BaseSceneB}{24.9}
\newcommand{\LearnedRoundOne}{58.3}

\newcommand{\RealBaseAverage}{20.0}
\newcommand{\RealStartAverage}{20.0}

\newcommand{\RealOursAverage}{76.7}

\newcommand{\DreamZeroBaseAverage}{27.2}
\newcommand{\DreamZeroOursAverage}{46.4}

\newcommand{\DreamZeroStartAverage}{28.5}
\newcommand{\VLMOnlyRoundFour}{43.7}
\newcommand{\SmallVLMRoundFour}{65.7}

\begin{document}

\abstract{
Improving robot policies on new tasks without collecting
additional expert demonstrations remains a central challenge
in robot learning.
World action models (WAMs) use broad video priors to jointly
predict future videos and actions, offering a potential source of supervision for adapting to new tasks.
However, generated videos may fail to depict task completion, and even visually successful videos may be paired with inconsistent actions that lead to execution failure.
We propose \textbf{EVO-WAM}, a framework that adapts WAMs to unseen tasks by learning from their own generated video-action trajectories, without executing candidate actions in an external environment.
First, we augment WAM training with state prediction and
anchored multi-frame context to enable complete autoregressive
rollouts without external execution feedback.
Second, we identify reliable training experience by selecting
task-completing prefixes with a vision-language model and
verifying their video-action consistency with an inverse
dynamics model.
Third, we iteratively train the WAM on verified prefixes and
generate new rollouts with the updated model.
On seven unseen RoboTwin 2.0 tasks, \method{}
increases average success rates from
\StartAverage\% to \OursAverage\% for Cosmos3 and from
\DreamZeroStartAverage\% to \DreamZeroOursAverage\% for DreamZero,
reaching approximately $2.5\times$ and $1.6\times$ their
initial success rates.
On three unseen long-horizon composite tasks in the real world,
it improves Cosmos3's average success rate
from \RealStartAverage\% to \RealOursAverage\%, a gain of
56.7 percentage points.
\textbf{Project Page:} {\urlstyle{tt}\url{https://evo-wam.github.io/}}.
}
\maketitle

\begingroup
\begin{figure}[H]
\centering
\includegraphics[width=\linewidth]{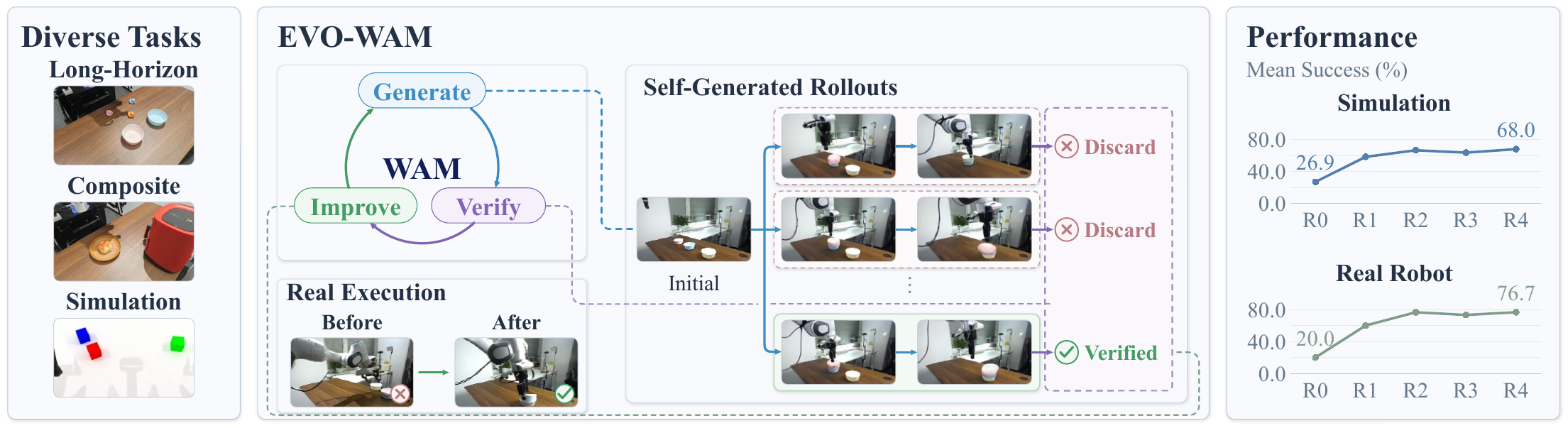}
\caption{\textbf{\method{} improves world action models through verified experience.}
Left: unseen tasks in simulation and long-horizon composite tasks in the real world.
Middle: the generate-verify-improve cycle, where learning from verified rollouts
improves task execution. Right: Cosmos3's performance gains over successive rounds
in simulation and the real world.}
\label{fig:rsi-teaser}
\end{figure}
\endgroup

\section{Introduction}
\label{sec:intro}
Adapting robot policies to unseen tasks without collecting additional expert demonstrations remains a central challenge in robot learning. Large-scale robot datasets~\citep{DBLP:conf/rss/KhazatskyP0BDKN24,DBLP:journals/corr/abs-2509-00576,DBLP:journals/corr/abs-2512-24653,DBLP:journals/corr/abs-2511-17441} have enabled increasingly general policies, but extending demonstration coverage to new tasks remains costly~\citep{DBLP:journals/corr/abs-2608-27550}.

To support such generalization, world action models (WAMs)~\citep{DBLP:conf/cvpr/BiTXWHLZFXRZLSM26,DBLP:journals/corr/abs-2603-22078,DBLP:conf/eccv/ChenLZWLYSLXWCYT26,wu2026zimablueevolvinggeneralizableworld} draw on broad video priors to jointly predict future videos and actions. Acquired through large-scale video pretraining~\citep{DBLP:journals/corr/abs-2606-02800,DBLP:journals/corr/abs-2503-20314}, these priors capture motion, physical interactions, and scene evolution, allowing video predictions to guide action generation and planning~\citep{DBLP:journals/corr/abs-2504-20995,DBLP:conf/iclr/KoMDST24,DBLP:journals/corr/abs-2602-20057}. Recent models, including DreamZero~\citep{DBLP:journals/corr/abs-2602-15922}, Cosmos3, and LingBot-VA~\citep{DBLP:journals/corr/abs-2601-21998}, can complete unseen tasks or make progress toward their goals in some trials, suggesting that video priors offer useful knowledge beyond the tasks covered by robot demonstrations. 

However, this potential does not guarantee successful adaptation
to unseen tasks. Specifically, as illustrated in
Figure~\ref{fig:case-ducks}, existing WAMs do not consistently generate videos depicting task completion.
Even under the same initial conditions, generated videos may
depict either success or failure. Moreover, the paired actions
may be inconsistent with the visually depicted behavior,
leading to execution failure even when the video depicts
success. This raises a question: \textbf{\textit{Can a WAM improve its performance on unseen tasks by identifying and learning from reliable trajectories within its own generated rollouts?}}

Existing methods have used separate world models to generate experience for policy improvement~\citep{DBLP:journals/corr/abs-2602-12063,DBLP:journals/corr/abs-2511-09515,DBLP:journals/corr/abs-2510-10125,jang2025dreamgenunlockinggeneralizationrobot,DBLP:journals/corr/abs-2608-08558,kim2026robocurateharnessingdiversityactionverified}. WAM-generated replay has also been used to preserve previously learned skills during continual learning~\citep{DBLP:journals/corr/abs-2606-27374}. However, these approaches still rely on experience from a separate world model or task demonstrations to learn new tasks, leaving the WAM’s own video priors unexploited as a source of supervision for unseen-task improvement.

Our key observation is that a WAM can improve on unseen tasks by learning from its own generated rollouts that depict successful task completion and preserve video-action consistency, without executing actions in an external environment. However, obtaining such rollouts raises two
challenges. First, the model must generate complete rollouts without receiving updated observations or robot states from an external environment. Second, it requires actions that are consistent with the generated video and can faithfully realize the depicted behavior during execution.

Motivated by this observation, we present \textbf{\method{}}, a framework for improving world action models through video-action verification, as shown in Fig.~\ref{fig:rsi-teaser} and ~\ref{fig:rsi-framework}. We augment WAM training with state prediction and anchored multi-frame context, enabling complete autoregressive continuation without external execution feedback. To identify reliable
rollouts, we use a vision-language model (VLM) to select task-completing prefixes and an
inverse dynamics model (IDM)~\citep{tian2025predictive} to verify their video-action consistency. We train the WAM on prefixes that pass both stages
and use the updated model to generate new candidates, iteratively
improving performance on unseen tasks.

We evaluate \method{} across two WAM backbones, Cosmos3~\citep{DBLP:journals/corr/abs-2606-02800} and DreamZero~\citep{DBLP:journals/corr/abs-2602-15922}, on seven RoboTwin 2.0~\citep{DBLP:journals/corr/abs-2506-18088} tasks unseen during base-model training, where self-training on verified trajectories increases average success rates from \StartAverage\% to \OursAverage\% and from \DreamZeroStartAverage\% to \DreamZeroOursAverage\%, respectively. We further evaluate \methodbackbone{Cosmos3} on three unseen long-horizon composite tasks in the real world, improving average success from \RealStartAverage\% to \RealOursAverage\%. Additionally, our study in Section~4.5 shows that verifying both task completion and video-action consistency is important for substantial gains from self-training.
This verification process is effective with both the large, highly capable Qwen3.8-Flash-Next~\citep{DBLP:journals/corr/abs-2608-30320} and the smaller, efficient Qwen3.5-27B~\citep{qwen3.5}. Our seen-task evaluation further shows that \methodbackbone{Cosmos3} largely preserves performance on seen tasks.

In summary, our contributions are threefold:
\begin{itemize}
    \item \textbf{A framework for improving WAMs with generated experience.}
    We introduce \method{}, which enables WAMs to generate autoregressive video-action rollouts and improve on unseen tasks through verification and iterative self-training, without additional expert demonstrations or action execution in an external environment during self-evolution.

    \item \textbf{Verification of task completion and video-action consistency.}
    We introduce a two-stage verification process in which a VLM identifies task-completing prefixes and an IDM assesses their video-action consistency, selecting generated rollouts for iterative self-training.

    \item \textbf{Generality across WAM backbones, tasks and VLMs.}
    We demonstrate improvements across Cosmos3 and DreamZero on unseen RoboTwin tasks and with Cosmos3 on real-world long-horizon composite tasks. It also shows robustness to VLM choice, sustaining performance gains with verifiers of different sizes.
    
\end{itemize}

\section{Preliminaries}
\label{sec:preliminaries}

\subsection{World Action Models}
\label{sec:wam-preliminaries}

World action models (WAMs) jointly predict future videos and robot actions conditioned
on visual observations, robot states, and task instructions~\citep{DBLP:journals/corr/abs-2602-15922,DBLP:journals/corr/abs-2601-21998}. At chunk $k$, a WAM with
parameters $\theta$ samples
\[
(\hat V_k,\hat A_k)\sim p_\theta^{\mathrm{VA}}(\,\cdot\mid h_k,\ell),
\]
where $h_k$ contains the visual context and robot state, $\ell$ is the task instruction,
and $\hat V_k$ and $\hat A_k$ denote the predicted video latents and actions. The decoded
video depicts anticipated task behavior, while the actions are intended to realize it
through execution.

Conventional vision-language-action (VLA) policies predict actions from observations
and task instructions~\citep{DBLP:conf/corl/ZitkovichYXXXXW23,DBLP:conf/corl/KimPKXB0RFSVKBT24}. Constructing new rollout
trajectories requires future observations from environment interaction or a separate
learned world model~\citep{DBLP:journals/corr/abs-2602-12063,DBLP:journals/corr/abs-2511-09515,DBLP:conf/icml/GaoZDZ025}. WAMs jointly predict the
visual frames and paired actions that can form training trajectories. This capability
gives WAMs the potential to learn from their own generated experience without external
execution feedback.

\subsection{Self-Training with Generated Rollouts}
\label{sec:rollout-self-training}

In closed-loop control, WAMs such as DreamZero refresh their visual
context with observations after action execution~\citep{DBLP:journals/corr/abs-2602-15922,DBLP:journals/corr/abs-2601-21998}.
Generating complete rollouts without this feedback requires predicted visual context
and robot states to condition subsequent chunks.

Moreover, even when such trajectories can be generated, they do not necessarily provide
reliable supervision. First, videos sampled from the same initial conditions and task
instruction may depict either task success or failure, so task completion cannot be
assumed from generation alone. Second, even when a video depicts success, its paired actions may be inconsistent
with the depicted behavior and fail to realize it during execution~\citep{DBLP:journals/corr/abs-2606-27374}. Training on rollouts
with either limitation may reinforce incomplete behaviors or actions that do not
achieve the imagined outcome. These limitations motivate assessing both visual task completion and video-action
consistency before using generated rollouts for self-training.

\section{Method}
\label{sec:method}
\begin{figure}[!htbp]
\centering
\includegraphics[width=\linewidth]{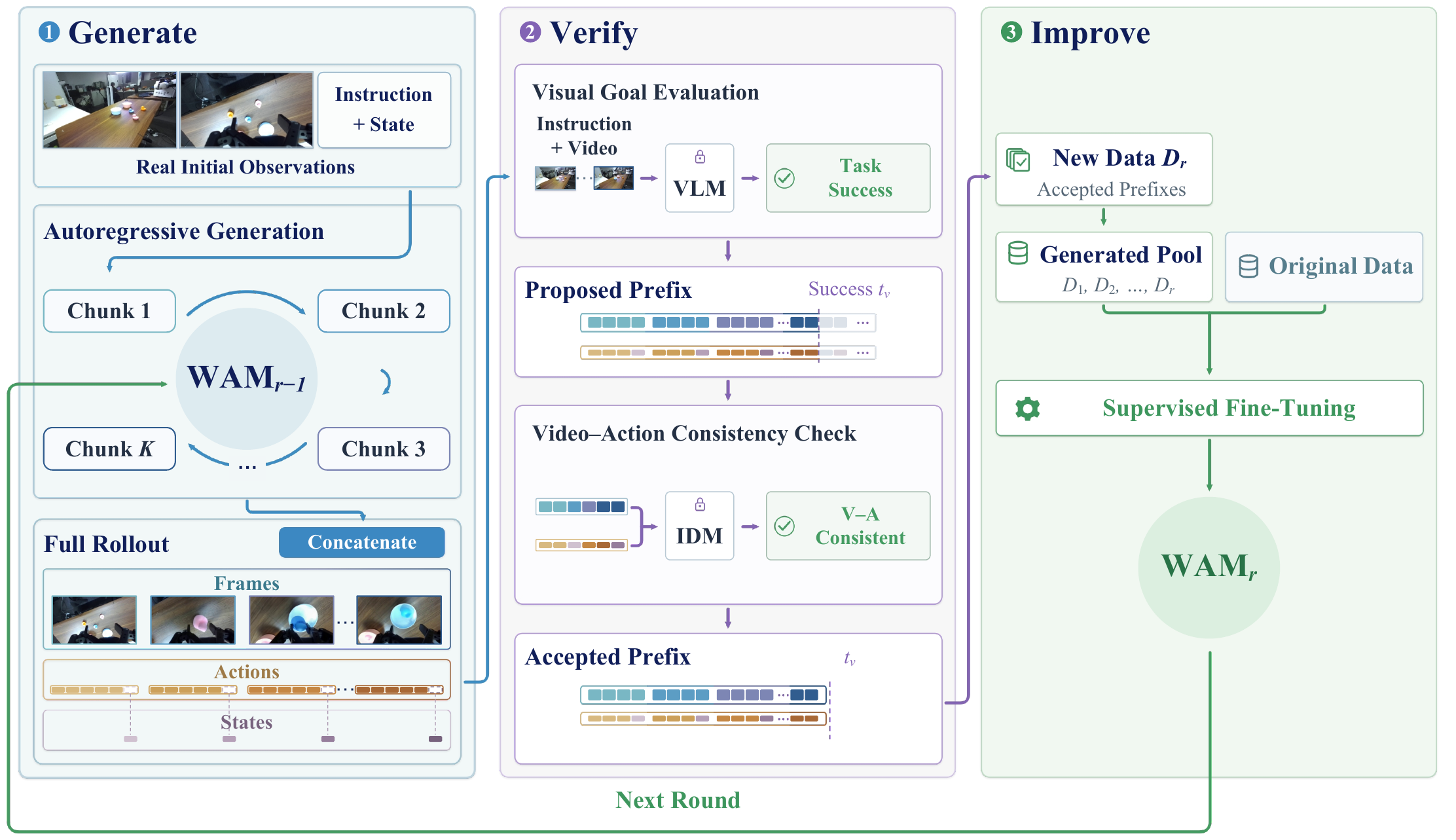}
\caption{\textbf{Overview of \method{}.} (1) Generate: $\mathrm{WAM}_{r-1}$
autoregressively generates video-action-state trajectories from an initial scene
and task instruction. (2) Verify: a VLM identifies task-completing prefixes, and an
IDM checks their video-action consistency. (3) Improve: verified prefixes accumulated
across rounds are combined with the original training data to obtain
$\mathrm{WAM}_r$, which generates candidates for the next round.}
\label{fig:rsi-framework}
\end{figure}

\paragraph{Overview.}
We present \method{}, a framework for improving WAMs on unseen tasks through
video-action verification, as shown in Fig.~\ref{fig:rsi-framework}. For each scene of an unseen task, we are
given an initial observation $o_0$, a robot state $s_0$, and a task instruction $\ell$.
Our goal is to improve task performance without additional expert demonstrations or action
execution in an external environment. We enable autoregressive rollouts through state
prediction and context augmentation (Sec.~\ref{sec:autoregressive-rollouts}), and select
reliable prefixes through video-action verification (Sec.~\ref{sec:video-action-verification}).
Starting from a base model $\mathrm{WAM}_0$, we repeat generation, verification, and training
over multiple rounds, updating $\mathrm{WAM}_{r-1}$ to $\mathrm{WAM}_r$ using verified data
at each round $r$ (Sec.~\ref{sec:recursive-update}).

\subsection{Enabling Autoregressive Rollouts}
\label{sec:autoregressive-rollouts}
\paragraph{State prediction and anchored multi-frame context.}
Autoregressive continuation requires the robot state and visual context for the next chunk.
We therefore train the WAM to predict the robot state at the end of each chunk, providing the robot
configuration needed for continuation. For visual conditioning, we initialize the rollout
with a single frame and retain it as an anchor alongside recent generated frames during
continuation. The recent frames provide motion history, while the anchor remains a
persistent reference for the objects and scene.

Let $\hat V_k$ and $\hat A_k$ denote the generated video latent and action blocks for chunk
$k$, and let $\hat s_k^{+}$ denote its predicted end state. Hats indicate model-generated
quantities. With $z_0$ denoting the latent representation of the initial observation $o_0$,
the conditioning context is
\begin{equation}
h_k=
\begin{cases}
(z_0,s_0), & k=1,\\[3pt]
\bigl(
z_0,\operatorname{Tail}_m(\hat V_{k-1}),
\hat s_{k-1}^{+}
\bigr), & k\geq 2,
\end{cases}
\label{eq:rollout-context}
\end{equation}
where $s_0$ is the initial state and $\operatorname{Tail}_m$ selects the last $m$ latent
frames of the video block. Both initialization and continuation modes are used
during base training and each self-training round.
\paragraph{Autoregressive trajectory generation.}
At self-training round $r$, let $\theta_{r-1}$ denote the parameters of $\mathrm{WAM}_{r-1}$. The model
generates each chunk conditioned on $h_k$ and the task instruction $\ell$:
\begin{equation}
(\hat V_k,\hat A_k,\hat s_k^{+})
\sim
p_{\theta_{r-1}}(\,\cdot\mid h_k,\ell).
\label{eq:chunk-generation}
\end{equation}
The generated video and predicted state then provide the context for the next chunk.
Repeating this process for a task-specific budget of $K$ chunks produces a candidate
trajectory $\hat\tau$, which includes the initial conditions $(o_0,s_0,\ell)$ and the
generated video-action-state sequence. These rollouts provide candidates for the
video-action verification described in Sec.~\ref{sec:video-action-verification}.

\subsection{Video-Action Verification}
\label{sec:video-action-verification}

\begin{figure}[!t]
\centering
\includegraphics[width=\linewidth]{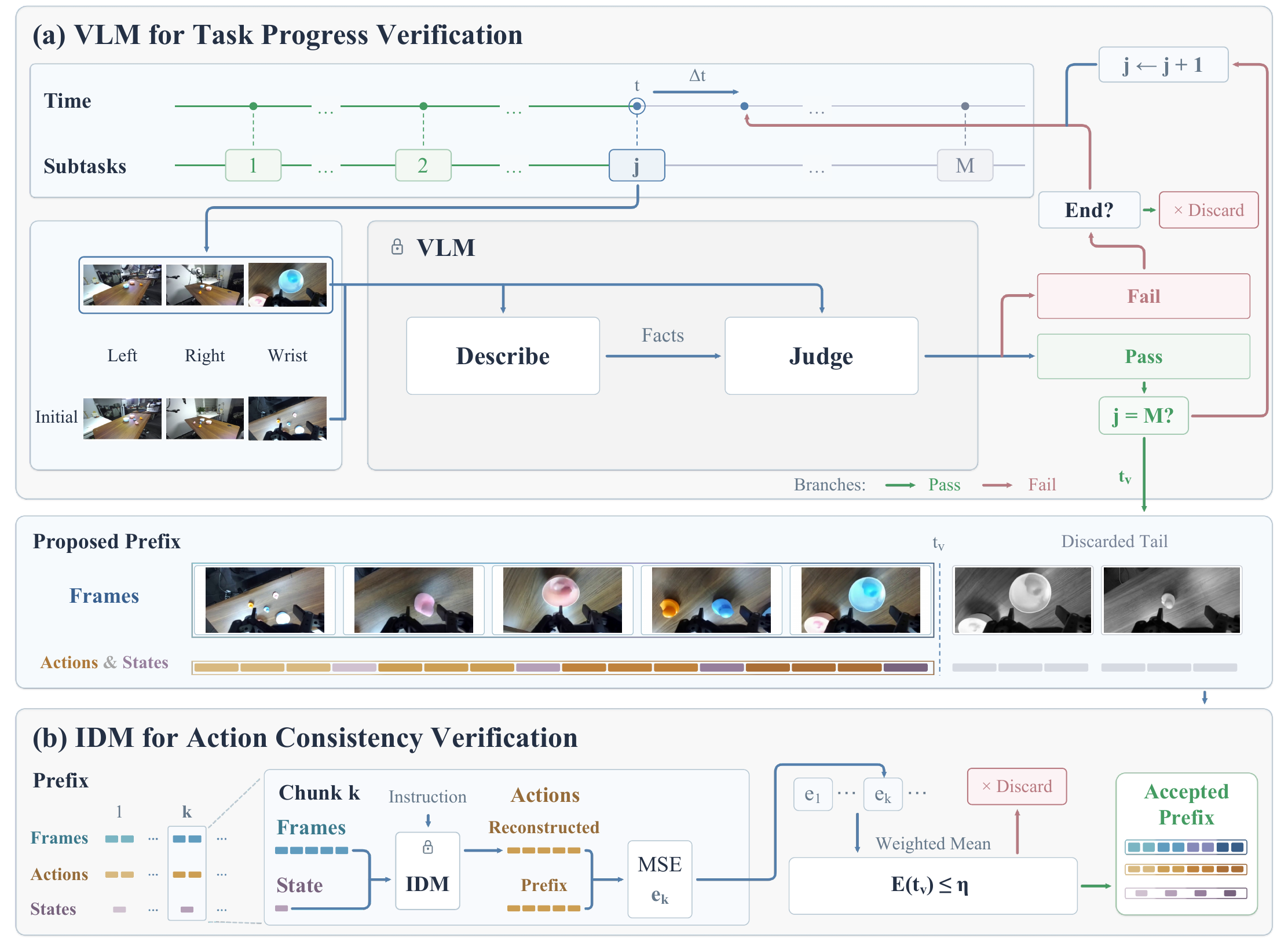}
\caption{\textbf{Video-action verification.} (a) A VLM checks subgoals one at a time
through description and judgment. (b) An IDM reconstructs actions from the generated videos and compares
them with WAM-generated actions at the fixed visual endpoint. Prefixes passing
this check undergo endpoint vote confirmation before entering training.}
\label{fig:rsi-verification}
\end{figure}

We select self-training prefixes in two stages, as shown in
Fig.~\ref{fig:rsi-verification}. A vision-language model (VLM) first identifies
task-completing prefixes. We then use an inverse dynamics model (IDM), which infers
actions from transitions between observations~\citep{DBLP:conf/nips/DuY0DN0SA23,DBLP:conf/icml/ZhouDCLYG24}, to reconstruct actions from the
generated videos. Comparing these reconstructions with the paired WAM-generated
actions provides a measure of video-action consistency.

\paragraph{Task-completion verification.}
We separate each VLM assessment into visual description and task judgment, grounding the
decision in explicit visual evidence. The VLM first describes the objects, their spatial
relations, and the robot configuration in the initial and generated observations from
multiple camera views. It then checks these descriptions against the same images, the task
instruction, and the active subgoal, using the initial scene as a reference. The judgment
evaluates goal satisfaction, required gripper release, object consistency, and robot
structural consistency. An assessment returns Accept only when all four verification checks are satisfied.

For both RoboTwin and real-robot tasks, we scan predefined time points chronologically
and check subgoals $(g_1,\ldots,g_M)$ in sequence ($M=1$ for a single goal).
We fix each $t_j$ at the first VLM Accept after $t_{j-1}$, then set $t_v=t_M$.
Once all endpoints are fixed, we check the prefix's video-action consistency.
Only if this check passes do we perform two additional description-judgment
assessments at each endpoint using the same observations and subgoal.
Each endpoint must receive at least two Accept judgments out of three.
A missing visual endpoint or a failed consistency or voting check discards the
candidate; verification does not resume at a later endpoint. Accepted prefixes
are retained through $t_v$. Appendix~\ref{app:task-completion-details}
formalizes this procedure in Algorithm~\ref{alg:prefix-verification}.

\paragraph{Video-action consistency verification.}
An IDM trained on recorded video-action pairs provides a reference for the actions
associated with depicted motion.
We adapt a pretrained video model into an action-only IDM conditioned on a video
window, its robot state, and the task instruction, with implementation
details in Appendix~\ref{app:idm-details}. We compare its reconstructions with
the WAM-generated actions in the same normalized action space, as shown in
Fig.~\ref{fig:rsi-verification}(b).

For a prefix ending at $t_v$, we summarize this discrepancy as $E(t_v)$, the mean squared
error across verification windows. The consistency check
passes if $E(t_v)\leq\eta$, where $\eta$ is fixed across self-training rounds for each
backbone and dataset. Calibration is described in Appendix~\ref{app:idm-details}.
Prefixes that pass both verification stages provide the
self-training data used in Sec.~\ref{sec:recursive-update}.
\subsection{Iterative Self-Training}
\label{sec:recursive-update}

As shown in Fig.~\ref{fig:rsi-framework}, we improve the WAM through iterative training on
prefixes retained after generation and verification. At round $r$, we combine the newly
verified data $\mathcal D_r$ with all retained data from earlier rounds,
$\mathcal D_1,\ldots,\mathcal D_{r-1}$, to preserve trajectory diversity and limit shifts
in the training distribution between updates. We mix these data with the original training
data $\mathcal D_{\mathrm{base}}$ and update the WAM by supervised fine-tuning:
\begin{equation}
\theta_r
\leftarrow
\operatorname{SFT}
\left(
\theta_{r-1};
\mathcal D_{\mathrm{base}},\mathcal D_1,\ldots,\mathcal D_r
\right).
\label{eq:recursive-update}
\end{equation}

The updated $\mathrm{WAM}_r$ generates candidates for the next round, continuing the cycle
of generation, verification, and training. Table~\ref{tab:recursive-improvement} reports
performance improvements over successive rounds.

\begin{figure}[!t]
\centering
\includegraphics[width=\linewidth]{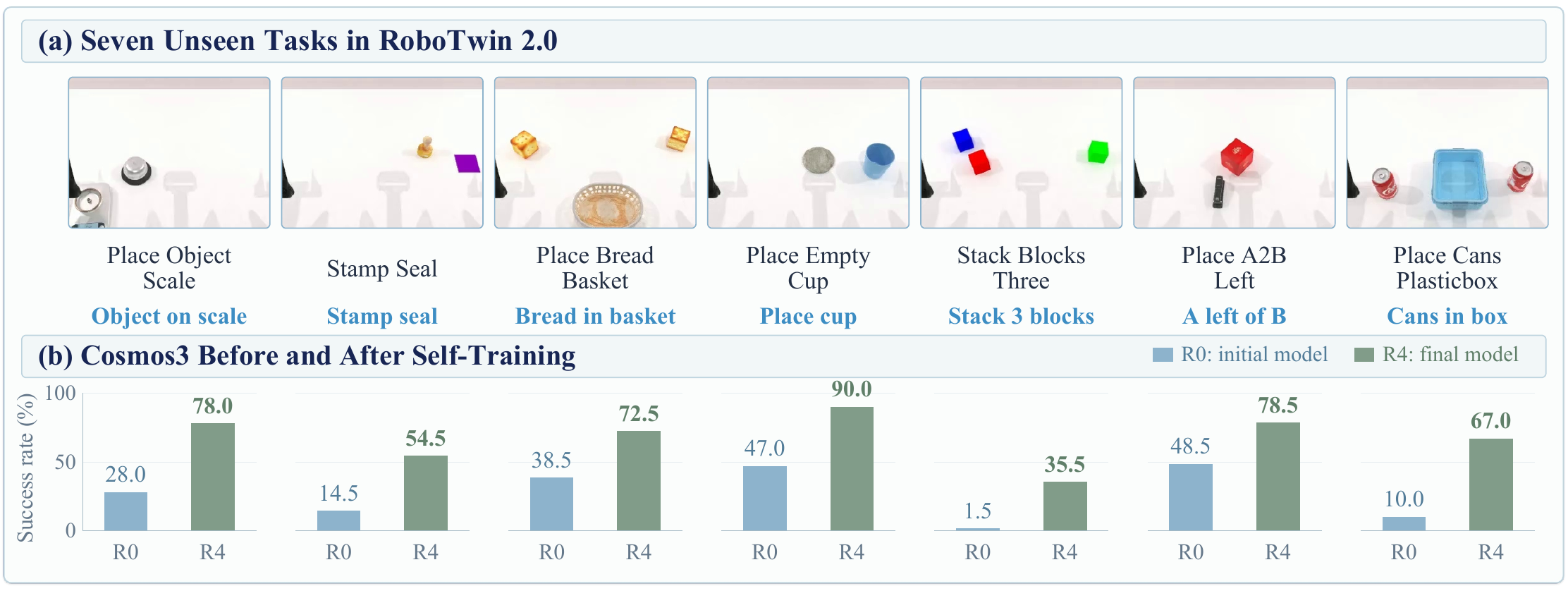}
\caption{\textbf{Unseen-task improvement in RoboTwin 2.0.}
(a) Example task scenes.
(b) Cosmos3's per-task success rates before self-training (Round 0) and after four rounds (Round 4).}
\label{fig:robotwin-task-performance}
\end{figure}

\begin{table}[!t]
\centering
\caption{\textbf{RoboTwin 2.0 results on unseen tasks.} Success rates (\%) on seven tasks unseen during base-model training. \methodbackbone{Cosmos3} achieves \OursAverage\% average success, compared with \BaseAverage\% for the strongest baseline Cosmos3. Baselines are trained for 34K steps; \method{} models start from
30K checkpoints and are evaluated at 34K steps. Bold and underlined values indicate the best and second-best results.}
\label{tab:robotwin-main}
\begingroup
\normalsize
\setlength{\tabcolsep}{0.8pt}
\renewcommand{\arraystretch}{1.16}
\begin{tabularx}{\linewidth}{l|>{\centering\arraybackslash}X>{\hsize=0.9\hsize\linewidth=\hsize\centering\arraybackslash}X>{\hsize=1.1\hsize\linewidth=\hsize\centering\arraybackslash}X*{4}{>{\centering\arraybackslash}X}|>{\centering\arraybackslash}p{4.3em}}
\toprule
\tablehead{Method} & \tablehead{\shortstack{Object\\on scale}} & \tablehead{\shortstack{Stamp\\seal}} & \tablehead{\shortstack{Bread\\in basket}} & \tablehead{\shortstack{Place\\cup}} & \tablehead{\shortstack{Stack\\3 blocks}} & \tablehead{\shortstack{A left\\of B}} & \tablehead{\shortstack{Cans\\in box}} & \tablehead{Average} \\
\midrule
\multicolumn{9}{c}{\textbf{\textit{Vision-Language-Action Models}}} \\
\midrule
$\pi_{0.5}$ & 17.0 & 14.0 & 14.0 & 44.0 & 0.0 & 29.0 & 0.0 & 16.9 \\
LingBot-VLA & 13.0 & 0.5 & 29.0 & 17.0 & 0.0 & 21.5 & 2.0 & 11.9 \\
StarVLA-OFT & 0.0 & 0.0 & 8.5 & 21.5 & 0.0 & 0.5 & 0.0 & 4.4 \\
\midrule
\multicolumn{9}{c}{\textbf{\textit{World Action Models}}} \\
\midrule
Fast-WAM & 3.5 & 1.0 & 22.5 & 31.5 & 0.0 & 0.0 & 0.0 & 8.4 \\
LingBot-VA & 7.0 & 5.5 & 8.0 & 29.5 & 0.0 & 23.0 & 0.0 & 10.4 \\
DreamZero & 26.5 & 25.5 & 39.5 & 25.0 & 0.0 & 59.5 & 14.5 & 27.2 \\
Cosmos3 & 24.0 & 22.5 & 40.5 & \underline{54.5} & \underline{5.0} & 55.5 & 19.0 & 31.6 \\
\midrule
\multicolumn{9}{c}{\textbf{\textit{World Action Models with \method{} (Ours)}}} \\
\midrule
\rowcolor{oursblue}
\textbf{\methodbackbone{DreamZero}} & \underline{61.5} & \underline{36.5} & \textbf{78.5} & 25.0 & 2.0 & \underline{73.5} & \underline{48.0} & \underline{46.4} \\
\rowcolor{oursblue}
\textbf{\methodbackbone{Cosmos3}} & \textbf{78.0} & \textbf{54.5} & \underline{72.5} & \textbf{90.0} & \textbf{35.5} & \textbf{78.5} & \textbf{67.0} & \textbf{68.0} \\
\bottomrule
\end{tabularx}
\endgroup
\end{table}
\section{Experiments}
\label{sec:experiments}

\subsection{Implementation}
\label{sec:implementation}

We apply \method{} to Cosmos3~\citep{DBLP:journals/corr/abs-2606-02800}
and DreamZero~\citep{DBLP:journals/corr/abs-2602-15922}, using subscripts to identify the backbone.
We use Qwen3.8-Flash-Next~\citep{DBLP:journals/corr/abs-2608-30320}
for task-completion assessment. IDM training details are provided
in Appendix~\ref{app:idm-details}. Self-training uses no additional
expert demonstrations or action execution in an external environment.

\subsection{Simulation Experiments}
\label{sec:simulation-results}

\paragraph{Setting and baselines.}
We use 43 RoboTwin 2.0 tasks~\citep{DBLP:journals/corr/abs-2506-18088}
for base-model training and the remaining seven for self-training
and evaluation, as shown in Figure~\ref{fig:robotwin-task-performance}.
We compare against the Cosmos3 and DreamZero baselines,
VLA models $\pi_{0.5}$~\citep{DBLP:journals/corr/abs-2504-16054},
LingBot-VLA~\citep{DBLP:journals/corr/abs-2601-18692}, and
StarVLA-OFT~\citep{DBLP:journals/corr/abs-2604-05014,DBLP:journals/corr/abs-2502-19645},
and WAMs Fast-WAM~\citep{DBLP:journals/corr/abs-2603-16666} and
LingBot-VA~\citep{DBLP:journals/corr/abs-2601-21998}.
All baselines are trained for 34K steps with a global batch
size of 256. The main evaluation uses the same scene configurations
used for self-generation, with 100 Clean and 100 Randomized
trials per task.

\paragraph{Self-training settings.}
Starting from the 30K-step Cosmos3 and DreamZero checkpoints,
we perform four rounds of self-training. Each round generates a
budget of 2,800 candidate rollouts, followed by 1K training updates
with a global batch size of 256. Verified prefixes are
accumulated across rounds. Detailed generation and training
configurations for both self-training and baselines are provided
in Appendix~\ref{app:configuration}.
\paragraph{Results.}
As shown in Table~\ref{tab:robotwin-main}, \methodbackbone{Cosmos3} and
\methodbackbone{DreamZero} achieve \OursAverage\% and \DreamZeroOursAverage\%
average success, compared with \BaseAverage\% and \DreamZeroBaseAverage\%
for their respective baselines. Cosmos3 is the strongest baseline,
while $\pi_{0.5}$ performs best among the VLA baselines.
Figure~\ref{fig:robotwin-task-performance} shows \methodbackbone{Cosmos3}'s
per-task improvements from Round 0 to Round 4.

Both backbones benefit, but their gains vary across tasks.
DreamZero's limited improvement on empty-cup placement and block stacking
may reflect constraints on the useful behaviors available in its generated candidates for subsequent policy improvement.
\subsection{Real-World Experiments}
\label{sec:real-robot-results}
\begin{figure}[!t]
\includegraphics[width=\linewidth]{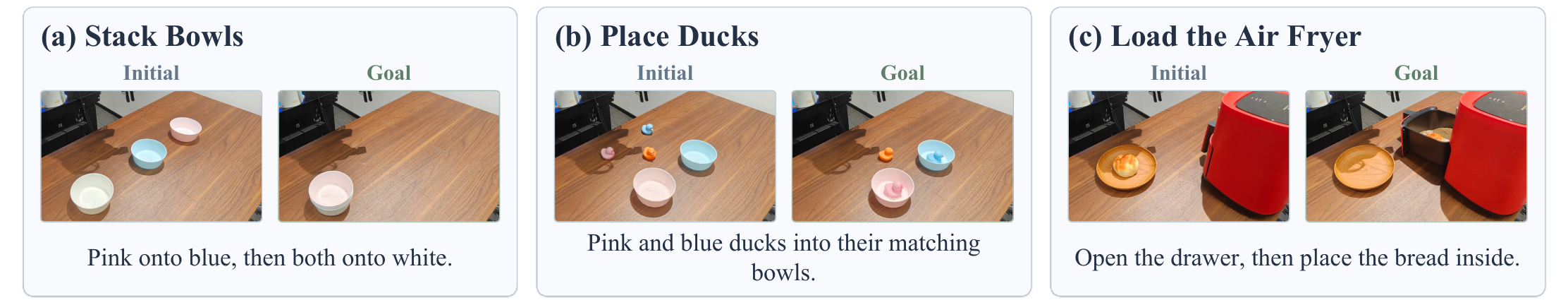}
\caption{\textbf{Real-robot tasks.} Initial and goal scenes for three unseen composite tasks.}
\label{fig:real-robot-tasks}
\end{figure}

\paragraph{Setting and baselines.}
We evaluate \methodbackbone{Cosmos3} on a Franka robot across three unseen
long-horizon and composite tasks, as shown in
Figure~\ref{fig:real-robot-tasks}. Stacking bowls tests
multi-stage manipulation and semantic understanding. Placing ducks
into matching bowls tests generalization to unseen objects
and target selection amid distractors. Loading an air fryer
tests the execution of composite subtasks, requiring the
drawer to be opened before bread is placed inside.

The Cosmos3 baseline is obtained by training the released
checkpoint for 31K additional steps on
DROID~\citep{DBLP:conf/rss/KhazatskyP0BDKN24} with state prediction
and context augmentation, as described in
Section~\ref{sec:autoregressive-rollouts}.
We also compare against unmodified public $\pi_{0.5}$ and
DreamZero checkpoints. Improvements use no additional expert demonstrations or feedback from executing candidate actions in the environment. Each policy is evaluated in ten trials per task.

\paragraph{Self-training settings.}
Starting from the 30K-step Cosmos3 checkpoint, we perform four rounds of
self-training. Each round uses an initial batch of 800 candidate
rollouts across the three tasks and performs 500 training updates
with a global batch size of 256. Verified prefixes are accumulated
across rounds. Additional details are reported in Appendix~\ref{app:configuration}.

\paragraph{Results.}
As shown in Table~\ref{tab:real-robot-main}, \methodbackbone{Cosmos3} achieves
\RealOursAverage\% average success, compared with \RealBaseAverage\%
for both Cosmos3 and DreamZero and 6.7\% for $\pi_{0.5}$.
It exceeds the strongest baseline on stacking bowls, placing
ducks, and loading the air fryer by 20, 50, and 80 percentage
points, respectively. These gains indicate improvements in multi-stage manipulation,
target selection amid distractors, and composite task completion.

\begin{figure}[!htbp]
\centering
\includegraphics[width=\linewidth,trim=0 132bp 0 132bp,clip]{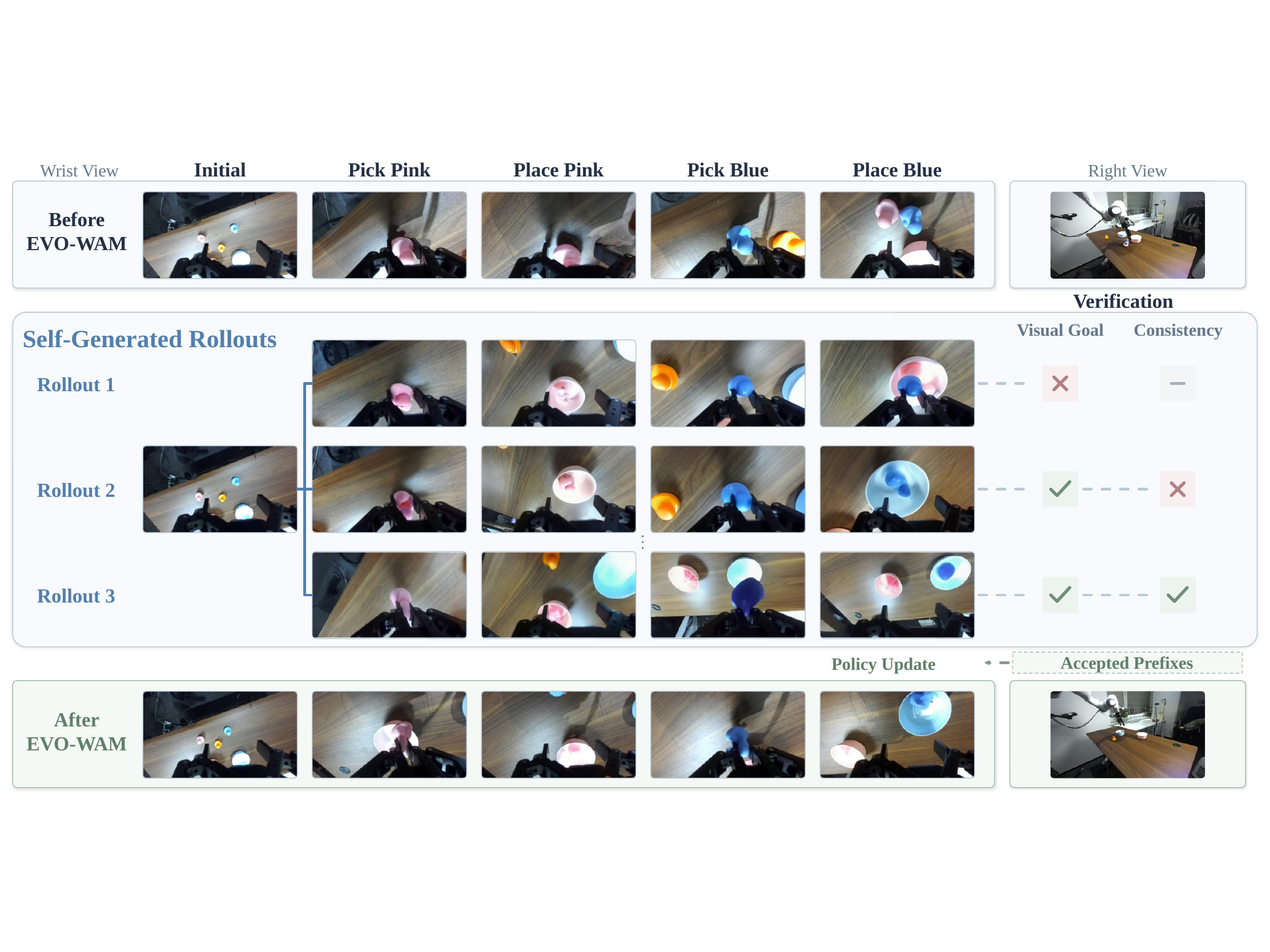}
\caption{\textbf{Placing two ducks.} Top and bottom show real executions before and
after \method{}. The middle shows generated candidates and their visual-goal and
video-action consistency checks; prefixes passing both checks are used to update
the policy.}
\label{fig:case-ducks}
\end{figure}

\paragraph{Qualitative example.}
In Figure~\ref{fig:case-ducks}, one candidate places the blue duck
in the pink bowl and fails visual-goal verification. Another
passes this check but fails video-action consistency verification.
After learning from prefixes that pass both checks, the policy
successfully completes both placements. Additional cases appear
in Appendix~\ref{app:real-robot-cases}.

\begin{table}[!htbp]
\centering
\caption{\textbf{Real-robot results.} Success rates (\%) on three unseen long-horizon composite tasks. \methodbackbone{Cosmos3} achieves \RealOursAverage\% average success, compared with \RealBaseAverage\% for Cosmos3. The Cosmos3 baseline and our Round 2 model are both evaluated at 31K steps. Bold and underlined values indicate the best and second-best results across all methods.}
\label{tab:real-robot-main}
\begingroup
\normalsize
\setlength{\tabcolsep}{5pt}
\renewcommand{\arraystretch}{1.13}
\begin{tabularx}{\linewidth}{l|>{\centering\arraybackslash}X>{\centering\arraybackslash}X>{\centering\arraybackslash}X>{\centering\arraybackslash}X}
\toprule
\tablehead{Model} & \tablehead{\shortstack{Stack\\Bowls}} & \tablehead{\shortstack{Place\\Ducks}} & \tablehead{\shortstack{Load the\\Air Fryer}} & \tablehead{Average} \\
\midrule
$\pi_{0.5}$ & 10.0 & \underline{10.0} & 0.0 & 6.7 \\
DreamZero & \underline{60.0} & 0.0 & 0.0 & \underline{20.0} \\
Cosmos3 & 40.0 & \underline{10.0} & \underline{10.0} & \underline{20.0} \\
\midrule
\rowcolor{oursblue}
\textbf{\methodbackbone{Cosmos3} (Ours)} & \textbf{80.0} & \textbf{60.0} & \textbf{90.0} & \textbf{76.7} \\
\bottomrule
\end{tabularx}
\endgroup
\end{table}

\subsection{Improvement over Multiple Rounds}
\label{sec:recursive-improvement}

Table~\ref{tab:recursive-improvement} and
Figure~\ref{fig:rsi-teaser} show that \methodbackbone{Cosmos3} gains most in
Round 1 (\StartAverage\% to \LearnedRoundOne\%). After four
rounds, \methodbackbone{Cosmos3} and \methodbackbone{DreamZero} reach \OursAverage\% and
\DreamZeroOursAverage\%, respectively. The Cosmos3 variant temporarily declines in Round 3
and recovers to 68.0\% in Round 4. On the real robot,
\methodbackbone{Cosmos3} reaches \RealOursAverage\% in both Rounds 2 and 4,
with 73.3\% in Round 3.
The large early gains suggest that useful supervision can already be extracted
from the initial model's generations. Later rounds bring smaller gains and
occasional regressions, showing that additional self-training does not always
improve performance.

\subsection{Ablation Studies}
\label{sec:ablations}

\paragraph{Action verification.}
To assess whether stricter action verification improves self-training,
we compare three criteria using the same starting policy and
Qwen3.8-Flash-Next, as shown in Table~\ref{tab:feedback}.
Controlled settings are detailed in Appendix~\ref{app:configuration}.
VLM only checks visual completion; VLM + IDM additionally checks
video-action consistency; VLM + Simulator retains prefixes whose
paired actions complete the task in simulation, providing
execution-verified supervision.

VLM + IDM outperforms VLM only in every round, reaching
\OursAverage\% versus \VLMOnlyRoundFour\% in Round 4.
This gap shows that visual completion alone is insufficient for
selecting effective action supervision. VLM + Simulator reaches
72.7\% in Round 4. Directly testing whether the actions complete the task may
explain its stronger performance. However, it requires a simulator of the target
task and scene. IDM-based verification also supports real-world tasks without
such simulators.
\begin{table}[!htbp]
\centering
\caption{\textbf{Improvement over successive self-training rounds.} Average success rates (\%) on seven unseen RoboTwin 2.0 tasks and three unseen real-world tasks. Parentheses show gains over Round 0 in percentage points (pp), computed before rounding. Round 0 denotes the 30K self-training initialization. Each simulation round adds 1K updates, and each real-world round adds 500 updates.}
\label{tab:recursive-improvement}
\begingroup
\normalsize
\setlength{\tabcolsep}{1.5pt}
\renewcommand{\arraystretch}{1.13}
\begin{tabularx}{\linewidth}{@{}l|c>{\centering\arraybackslash}X>{\centering\arraybackslash}X>{\centering\arraybackslash}X>{\centering\arraybackslash}X@{}}
\toprule
\tablehead{Model} & \tablehead{Round 0} & \tablehead{Round 1} & \tablehead{Round 2} & \tablehead{Round 3} & \tablehead{Round 4} \\
\midrule
\multicolumn{6}{c}{\textbf{\textit{RoboTwin}}} \\
\midrule
\textbf{\methodbackbone{DreamZero} (Ours)} & 28.5 & \mbox{36.5 {\normalsize (+8.0)}} & \mbox{42.3 {\normalsize (+13.8)}} & \mbox{45.1 {\normalsize (+16.6)}} & \textbf{\mbox{46.4 {\normalsize (+17.9)}}} \\
\textbf{\methodbackbone{Cosmos3} (Ours)} & 26.9 & \mbox{58.3 {\normalsize (+31.4)}} & \mbox{66.6 {\normalsize (+39.8)}} & \mbox{63.6 {\normalsize (+36.7)}} & \textbf{\mbox{68.0 {\normalsize (+41.1)}}} \\
\midrule
\multicolumn{6}{c}{\textbf{\textit{Real world}}} \\
\midrule
\textbf{\methodbackbone{Cosmos3} (Ours)} & 20.0 & \mbox{60.0 {\normalsize (+40.0)}} & \textbf{\mbox{76.7 {\normalsize (+56.7)}}} & \mbox{73.3 {\normalsize (+53.3)}} & \textbf{\mbox{76.7 {\normalsize (+56.7)}}} \\
\bottomrule
\end{tabularx}
\endgroup
\end{table}
\begin{table}[!htbp]
\centering
\caption{\textbf{Effect of verification on self-training.} Success rates (\%) on unseen RoboTwin 2.0 tasks across rounds, comparing action verification criteria and task-completion VLMs. }
\label{tab:feedback}
\begingroup
\normalsize
\setlength{\tabcolsep}{3pt}
\renewcommand{\arraystretch}{1.13}
\begin{tabularx}{\linewidth}{@{}ll|c>{\centering\arraybackslash}X>{\centering\arraybackslash}X>{\centering\arraybackslash}X>{\centering\arraybackslash}X@{}}
\toprule
\tablehead{Verification} & \tablehead{VLM} & \tablehead{R0} & \tablehead{R1} & \tablehead{R2} & \tablehead{R3} & \tablehead{R4} \\
\midrule
VLM only & Qwen3.8-Flash-Next & 26.9 & \mbox{37.9 {\normalsize (+11.1)}} & \mbox{42.7 {\normalsize (+15.9)}} & \mbox{39.6 {\normalsize (+12.7)}} & \mbox{43.7 {\normalsize (+16.9)}} \\
\rowcolor{oursblue}
\textbf{VLM + IDM (Ours)} & Qwen3.8-Flash-Next & 26.9 & \mbox{58.3 {\normalsize (+31.4)}} & \mbox{66.6 {\normalsize (+39.8)}} & \mbox{63.6 {\normalsize (+36.7)}} & \mbox{68.0 {\normalsize (+41.1)}} \\
VLM + Simulator & Qwen3.8-Flash-Next & 26.9 & \mbox{66.4 {\normalsize (+39.5)}} & \mbox{69.3 {\normalsize (+42.4)}} & \mbox{73.2 {\normalsize (+46.4)}} & \mbox{72.7 {\normalsize (+45.9)}} \\
VLM + IDM & Qwen3.5-27B & 26.9 & \mbox{52.7 {\normalsize (+25.9)}} & \mbox{57.8 {\normalsize (+30.9)}} & \mbox{62.7 {\normalsize (+35.9)}} & \mbox{65.7 {\normalsize (+38.9)}} \\
\bottomrule
\end{tabularx}
\endgroup
\end{table}

\paragraph{VLM sensitivity.}
To examine the effect of VLM selection, we use
Qwen3.5-27B~\citep{qwen3.5} for task-completion assessment
while retaining IDM-based action verification. As shown in
Table~\ref{tab:feedback}, using the smaller, less capable Qwen3.5-27B still yields
\SmallVLMRoundFour\% success in Round 4, compared with \OursAverage\% using
Qwen3.8-Flash-Next. Substantial gains with both VLMs suggest that \method{}
is robust to the choice of VLM, though stronger VLMs yield better performance.

\paragraph{Generalization to new scenes.}
The results in Table~\ref{tab:robotwin-main} show gains on the scene configurations
used for self-imagination and self-training. To further evaluate generalization,
we randomly sample 100 new scenes per task under each condition
and evaluate the updated model without further adaptation. As shown in
Table~\ref{tab:generalization-retention}, \methodbackbone{Cosmos3} achieves
\OursSceneB\% success, compared with \BaseSceneB\% for the baseline.
This improvement suggests that the learned behaviors extend beyond the initial
configurations used to generate training data. Details are in
Appendix~\ref{app:scene-set-details}.

\paragraph{Seen-task retention.}
On the 43 seen tasks with 50 Clean and 50 Randomized trials each,
\methodbackbone{Cosmos3} at Round 4 achieves \SeenFinal\% success, compared with
\SeenBase\% for the 34K Cosmos3 baseline (Table~\ref{tab:generalization-retention}).
These results show that it substantially improves unseen-task
performance while causing only a small decrease on seen tasks.

\begingroup
\begin{table}[!htbp]
\centering
\caption{\textbf{Generalization and retention (\%) of EVO-WAM.}}
\label{tab:generalization-retention}
\begingroup
\normalsize
\setlength{\tabcolsep}{8pt}
\renewcommand{\arraystretch}{1.13}
\begin{tabular}{lc>{\columncolor{oursblue}}c}
\toprule
Evaluation & Cosmos3 & \methodbackbone{Cosmos3} \\
\midrule
New-scene generalization & \BaseSceneB & \OursSceneB \\
Seen-task retention & \SeenBase & \SeenFinal \\
\bottomrule
\end{tabular}
\endgroup
\end{table}
\endgroup
\FloatBarrier
\section{Conclusion}
\label{sec:conclusion}
We presented \method{}, a framework that enables world action models to improve on tasks unseen during base-model training by learning from their own generated experience. By enabling autoregressive rollouts and verifying task completion and video-action consistency, the framework turns model predictions into self-training data without additional expert demonstrations or external action execution during self-training. Evaluations with Cosmos3 and DreamZero on RoboTwin 2.0 demonstrate improvements across both WAM backbones, while real-world evaluations with Cosmos3 show gains on long-horizon composite tasks. These findings point to a path toward self-improving robot policies, where video-action verification enables WAMs to transform their own predictions into experience for adapting to new tasks.
\bibliographystyle{evo_wam}
\bibliography{evo_wam}

\clearpage
\appendix
\section{Related Work}
\label{sec:related-work}

\paragraph{Learning from Self-Generated Data.}
Reasoning models use feedback to learn beyond expert demonstrations. STaR iteratively trains on model-generated rationales that lead to correct answers~\citep{DBLP:conf/nips/ZelikmanWMG22}. DeepSeek-R1~\citep{DBLP:journals/nature/GuoYZSWZXZMBZY025} develops reasoning through reinforcement learning with verifiable rewards. Absolute Zero~\citep{DBLP:conf/nips/ZhaoWWXYLWWZH25} generates and solves executable tasks, using a code executor to validate them, while SEAL~\citep{DBLP:conf/nips/ZweigerPGKA25} generates fine-tuning data and update directives and evaluates the resulting adaptation.
These approaches highlight the role of verification in learning from generated material. For embodied trajectories, verification must account for both task completion and consistency between visual outcomes and actions.

\paragraph{Self-Improvement in Robot Learning.}
DreamerV3~\citep{DBLP:journals/nature/HafnerPBL25} learns a world model from environment interaction and improves its policy using imagined trajectories.
Robot policies can improve through deployment experience. RECAP~\citep{DBLP:journals/corr/abs-2511-14759} combines autonomous rollouts, reward feedback, and human corrections to train $\pi^{*}_{0.6}$, while SILVR~\citep{luo2026selfimprovingloopsvisualrobotic} uses execution experience to improve visual planning on unseen tasks.
WMPO~\citep{DBLP:journals/corr/abs-2511-09515} and RISE~\citep{DBLP:journals/corr/abs-2602-11075} use learned world models to optimize policies through imagined interactions and reward or value feedback. Cosmos Policy~\citep{DBLP:journals/corr/abs-2601-16163} refines future-state and value prediction using rollout experience to improve model-based planning.
Our framework uses a WAM's own joint video-action predictions as supervision for iterative policy updates, without external action execution during adaptation.

\paragraph{Verifying Model-Generated Experience.}
GigaWorld-1~\citep{DBLP:journals/corr/abs-2607-02642} uses an action-conditioned world model for robot policy evaluation.
World Action Verifier~\citep{DBLP:journals/corr/abs-2604-01985} uses forward-inverse disagreement to
select informative environment interactions for updating a world model.
Our verification instead selects visually successful, video-action-consistent
prefixes from the WAM's own rollouts as training experience, without executing candidate actions in an external environment.

\paragraph{World Action Models and Unseen-Task Adaptation.}
For VLAs, VLAct~\citep{DBLP:journals/corr/abs-2608-27550} studies representation-centric continued
pretraining to improve transfer across tasks and embodiments.
UniPi~\citep{DBLP:conf/nips/DuY0DN0SA23} and RoboDreamer~\citep{DBLP:conf/icml/ZhouDCLYG24} generate video plans and recover actions through inverse dynamics. GR-1~\citep{DBLP:conf/iclr/WuJCCXLLLK24} jointly predicts images and actions after video pretraining, while Video Prediction Policy~\citep{DBLP:conf/icml/HuGWCWZSL025} conditions an action decoder on predictive video features.
DreamGen~\citep{jang2025dreamgenunlockinggeneralizationrobot} trains policies on video-generated trajectories
with recovered pseudo-actions. Vid2WAM~\citep{DBLP:journals/corr/abs-2608-08558} distills an external
video teacher into a WAM using generated futures and IDM-inferred actions.
In our framework, the WAM generates both training targets, and the IDM checks their consistency.
LingBot-VA~\citep{DBLP:journals/corr/abs-2601-21998} and DreamZero~\citep{DBLP:journals/corr/abs-2602-15922} jointly generate videos and actions, and Cosmos~3~\citep{DBLP:journals/corr/abs-2606-02800} supports world-action generation within an omnimodal backbone.
Zero-WAM~\citep{DBLP:journals/corr/abs-2608-26103} studies unseen-task execution with human video guidance, whereas ReGen~\citep{DBLP:journals/corr/abs-2606-27374} uses WAM-generated trajectories of previously learned tasks to mitigate forgetting during continual learning.
We instead generate and verify trajectories for tasks unseen during base-model training, then learn from them over successive rounds of self-improvement.

\section{Additional Experiments and Evaluation Details}
\label{app:experiments}

\subsection{Data Accumulation}
\label{app:data-accumulation}

\begin{figure}[!htbp]
\centering
\includegraphics[width=0.88\linewidth]{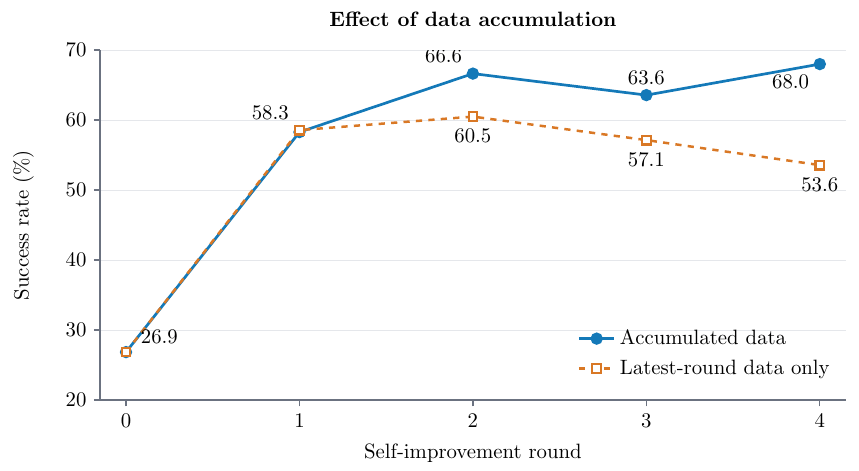}
\caption{\textbf{Effect of data accumulation.} Success in adaptation scenes across self-improvement
rounds under the two configurations described above.}
\label{fig:data-accumulation}
\end{figure}

Figure~\ref{fig:data-accumulation} and Table~\ref{tab:data-accumulation} report the
latest-round and accumulated-data series through Round 4. The accumulated-data results
are the \methodbackbone{Cosmos3} sequence from Table~\ref{tab:recursive-improvement}.
Both series use IDMs trained on the 43 seen RoboTwin tasks.

\begin{table}[!htbp]
\centering
\caption{\textbf{Effect of data accumulation.} Success rates (\%) corresponding to Figure~\ref{fig:data-accumulation}, with 1,400 trials per round. Accumulated data reproduces Table~\ref{tab:recursive-improvement}. Both series use IDMs trained on the 43 seen RoboTwin tasks.}
\label{tab:data-accumulation}
\begingroup
\normalsize
\setlength{\tabcolsep}{5pt}
\renewcommand{\arraystretch}{1.13}
\begin{tabular}{l|ccccc}
\toprule
\tablehead{Training data} & \tablehead{Round 0} & \tablehead{Round 1} & \tablehead{Round 2} & \tablehead{Round 3} & \tablehead{Round 4} \\
\midrule
Latest-round data only & 26.9 & 58.6 & 60.5 & 57.1 & 53.6 \\
Accumulated data & 26.9 & 58.3 & 66.6 & 63.6 & 68.0 \\
\bottomrule
\end{tabular}
\endgroup
\end{table}

\subsection{Per-Task Evaluation Across Scene Sets}
\label{app:scene-set-details}

The new-scene comparison in Table~\ref{tab:generalization-retention} uses the
34K Cosmos3 baseline from Table~\ref{tab:robotwin-main} and \methodbackbone{Cosmos3} at 34K steps,
with success rates of \BaseSceneB\% and \OursSceneB\%, respectively.
Both models are evaluated on the same new-scene test set, with 100 Clean and
100 Randomized trials per task (1,400 in total), using identical inference
settings and success criteria.
The 34K baseline is a separate comparison model; self-training starts from the
30K initialization described in Section~\ref{sec:simulation-results}.

Table~\ref{tab:scene-set-details} breaks down \methodbackbone{Cosmos3}'s Round 4 performance
in adaptation and new scenes. In new scenes, success reaches 82.0\% for placing bread in a basket
and 47.0\% for stacking three blocks, compared with 72.5\% and 35.5\% in adaptation scenes.

\begin{table}[!htbp]
\centering
\caption{\textbf{Per-task evaluation across scene sets.} Success rates (\%) of the final \methodbackbone{Cosmos3} model at 34K steps. Adapt. and New denote adaptation and new scenes; each set has 100 Clean and 100 Randomized trials per task (1,400 total). Average pools both conditions.}
\label{tab:scene-set-details}
\begingroup
\normalsize
\setlength{\tabcolsep}{3pt}
\renewcommand{\arraystretch}{1.18}
\begin{tabularx}{\linewidth}{l|*{4}{>{\centering\arraybackslash}X}|*{2}{>{\columncolor{oursblue}\centering\arraybackslash}X}}
\toprule
 & \multicolumn{2}{c}{\textbf{\textit{Clean}}} & \multicolumn{2}{c}{\textbf{\textit{Randomized}}} & \multicolumn{2}{>{\columncolor{oursblue}}c}{\textbf{\textit{Average}}} \\
\cmidrule(lr){2-3}\cmidrule(lr){4-5}\cmidrule(lr){6-7}
\tablehead{Task} & \tablehead{Adapt.} & \tablehead{New} & \tablehead{Adapt.} & \tablehead{New} & \tablehead{Adapt.} & \tablehead{New} \\
\midrule
Place object on scale & 79.0 & 84.0 & 77.0 & 75.0 & 78.0 & 79.5 \\
Stamp seal & 48.0 & 49.0 & 61.0 & 59.0 & 54.5 & 54.0 \\
Place bread in basket & 74.0 & 79.0 & 71.0 & 85.0 & 72.5 & 82.0 \\
Place empty cup & 93.0 & 89.0 & 87.0 & 86.0 & 90.0 & 87.5 \\
Stack three blocks & 34.0 & 49.0 & 37.0 & 45.0 & 35.5 & 47.0 \\
Place A to the left of B & 81.0 & 78.0 & 76.0 & 75.0 & 78.5 & 76.5 \\
Place cans in box & 65.0 & 67.0 & 69.0 & 65.0 & 67.0 & 66.0 \\
\midrule
\rowcolor{oursblue}
\textbf{Overall (Ours)} & \textbf{67.7} & \textbf{70.7} & \textbf{68.3} & \textbf{70.0} & \textbf{68.0} & \textbf{70.4} \\
\bottomrule
\end{tabularx}
\endgroup
\end{table}

\subsection{Real-Robot Tasks and Scene Configurations}
\label{app:real-robot}

\begin{figure}[!htbp]
\centering
\includegraphics[width=0.68\linewidth]{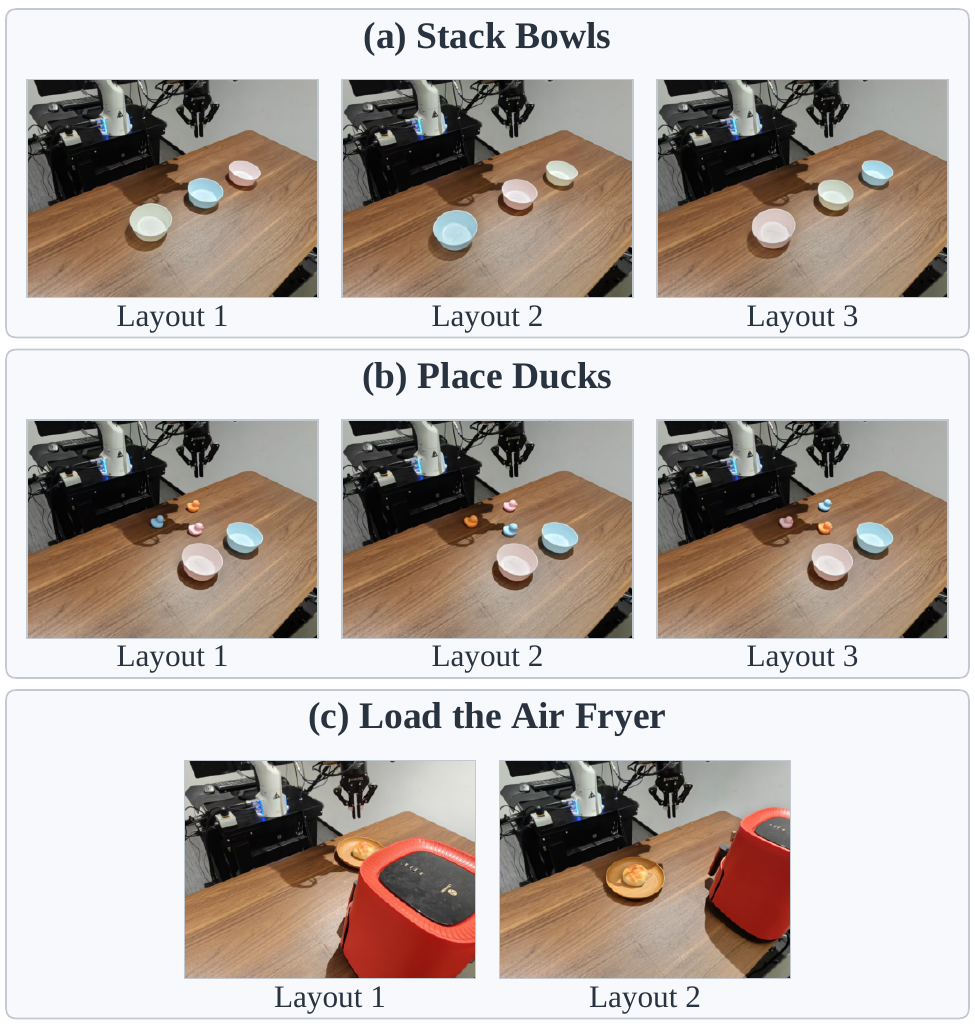}
\caption{\textbf{Initial configurations for physical evaluation.} Three layouts
for Stack Bowls, three for Place Ducks, and two for Load the Air Fryer. Each task
has ten trials per evaluated policy: 3/3/4 across the three-layout tasks and 5/5
across the two air-fryer layouts.}
\label{fig:real-robot-scenes}
\end{figure}

We evaluate each policy on ten physical trials per task. Stack Bowls and Place
Ducks each use three initial layouts, with three, three, and four trials. Load the
Air Fryer uses two layouts with five trials each. Figure~\ref{fig:real-robot-scenes}
shows the object arrangements. The instruction and target objects are fixed within
each task; their initial positions vary across layouts. Task success is the number of successful trials divided by ten.

\paragraph{Stack Bowls.}
The instruction is: ``Stack the pink bowl on the blue bowl, then lift both together
onto the white bowl.'' The three layouts permute the initial positions of the bowls.
Completion requires the white bowl to support the blue bowl, the blue bowl to
support the pink bowl, and the gripper to release the stack.

\paragraph{Place Ducks.}
The robot must place the pink toy duck in the pink bowl and then the blue toy duck
in the blue bowl. A third duck serves as a distractor. The three layouts change
the ducks' positions while retaining the two target bowls. Completion requires
both target ducks to be released into their corresponding bowls.

\paragraph{Load the Air Fryer.}
The robot must pull the red handle to open the drawer, pick up the bread from its
plate, and release it inside the drawer. The two layouts place the bread on
opposite sides of the air fryer.

Table~\ref{tab:real-robot-details} reports per-layout success counts for the baselines and all four self-training rounds;
Table~\ref{tab:real-robot-main} reports the task-level comparison, and
Table~\ref{tab:recursive-improvement} summarizes average success rates across
self-training rounds.
\begin{table}[!htbp]
\centering
\caption{\textbf{Real-robot results by layout.} Successful trials over attempts for the models in Table~\ref{tab:real-robot-main}. Layout numbers match Figure~\ref{fig:real-robot-scenes}. The Cosmos3 baseline is evaluated at 31K steps. R1--R4 denote the self-training rounds; Table~\ref{tab:real-robot-main} uses R2.}
\label{tab:real-robot-details}
\begingroup
\normalsize
\setlength{\tabcolsep}{2pt}
\renewcommand{\arraystretch}{1.13}
\begin{tabularx}{\linewidth}{l|>{\centering\arraybackslash}X>{\centering\arraybackslash}X>{\centering\arraybackslash}X>{\centering\arraybackslash}X>{\centering\arraybackslash}X>{\centering\arraybackslash}X>{\centering\arraybackslash}X}
\toprule
 & \multicolumn{1}{c}{\textit{VLA}} & \multicolumn{2}{c}{\textit{WAM}} & \multicolumn{4}{c}{\cellcolor{oursblue}\shortstack{\textbf{\methodbackbone{Cosmos3}}\\\textbf{(Ours)}}} \\
\cmidrule(lr){2-2}\cmidrule(lr){3-4}\cmidrule(lr){5-8}
\tablehead{Layout} & \tablehead{$\pi_{0.5}$} & \tablehead{DreamZero} & \tablehead{Cosmos3} & \tablehead{R1} & \tablehead{R2} & \tablehead{R3} & \tablehead{R4} \\
\midrule
\multicolumn{8}{c}{\textbf{\textit{Stack Bowls}}} \\
\midrule
Layout 1 & 0/3 & 3/3 & 1/3 & \cellcolor{oursblue}3/3 & \cellcolor{oursblue}3/3 & \cellcolor{oursblue}3/3 & \cellcolor{oursblue}3/3 \\
Layout 2 & 0/3 & 3/3 & 2/3 & \cellcolor{oursblue}3/3 & \cellcolor{oursblue}3/3 & \cellcolor{oursblue}2/3 & \cellcolor{oursblue}2/3 \\
Layout 3 & 1/4 & 0/4 & 1/4 & \cellcolor{oursblue}1/4 & \cellcolor{oursblue}2/4 & \cellcolor{oursblue}2/4 & \cellcolor{oursblue}3/4 \\
\cmidrule(lr){1-8}
\textbf{Total} & \textbf{1/10} & \textbf{6/10} & \textbf{4/10} & \cellcolor{oursblue}\textbf{7/10} & \cellcolor{oursblue}\textbf{8/10} & \cellcolor{oursblue}\textbf{7/10} & \cellcolor{oursblue}\textbf{8/10} \\
\midrule
\multicolumn{8}{c}{\textbf{\textit{Place Ducks}}} \\
\midrule
Layout 1 & 0/3 & 0/3 & 0/3 & \cellcolor{oursblue}0/3 & \cellcolor{oursblue}0/3 & \cellcolor{oursblue}0/3 & \cellcolor{oursblue}0/3 \\
Layout 2 & 1/3 & 0/3 & 1/3 & \cellcolor{oursblue}1/3 & \cellcolor{oursblue}3/3 & \cellcolor{oursblue}2/3 & \cellcolor{oursblue}2/3 \\
Layout 3 & 0/4 & 0/4 & 0/4 & \cellcolor{oursblue}1/4 & \cellcolor{oursblue}3/4 & \cellcolor{oursblue}4/4 & \cellcolor{oursblue}4/4 \\
\cmidrule(lr){1-8}
\textbf{Total} & \textbf{1/10} & \textbf{0/10} & \textbf{1/10} & \cellcolor{oursblue}\textbf{2/10} & \cellcolor{oursblue}\textbf{6/10} & \cellcolor{oursblue}\textbf{6/10} & \cellcolor{oursblue}\textbf{6/10} \\
\midrule
\multicolumn{8}{c}{\textbf{\textit{Load the Air Fryer}}} \\
\midrule
Layout 1 & 0/5 & 0/5 & 0/5 & \cellcolor{oursblue}5/5 & \cellcolor{oursblue}4/5 & \cellcolor{oursblue}4/5 & \cellcolor{oursblue}5/5 \\
Layout 2 & 0/5 & 0/5 & 1/5 & \cellcolor{oursblue}4/5 & \cellcolor{oursblue}5/5 & \cellcolor{oursblue}5/5 & \cellcolor{oursblue}4/5 \\
\cmidrule(lr){1-8}
\textbf{Total} & \textbf{0/10} & \textbf{0/10} & \textbf{1/10} & \cellcolor{oursblue}\textbf{9/10} & \cellcolor{oursblue}\textbf{9/10} & \cellcolor{oursblue}\textbf{9/10} & \cellcolor{oursblue}\textbf{9/10} \\
\bottomrule
\end{tabularx}
\endgroup
\end{table}

\subsection{Additional Real-Robot Case Studies}
\label{app:real-robot-cases}

\begin{figure}[!htbp]
\centering
\includegraphics[width=\linewidth,trim=0 132bp 0 132bp,clip]{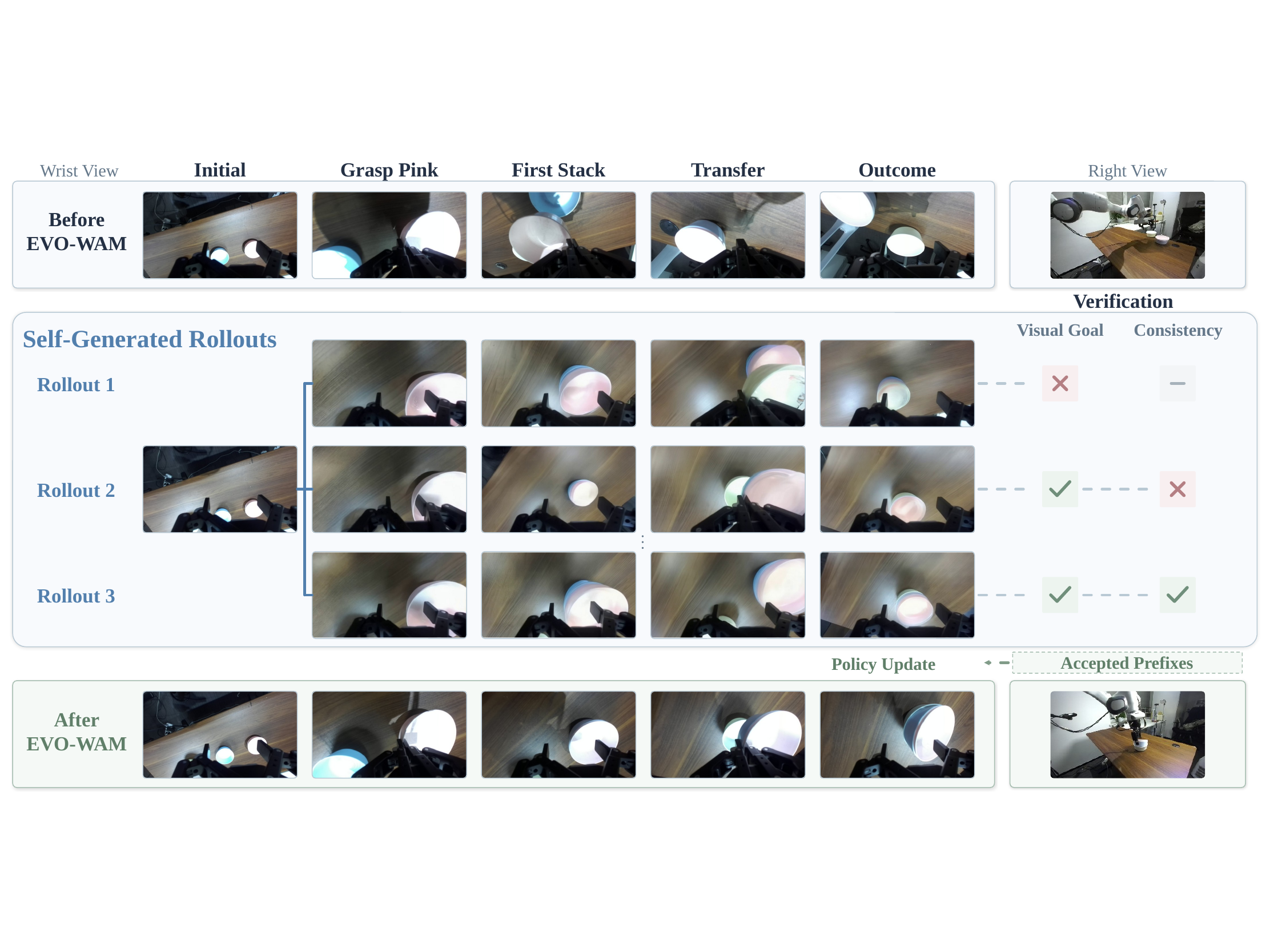}
\caption{\textbf{Stacking three bowls.} The instruction is: ``Stack the pink bowl
on the blue bowl, then lift both together onto the white bowl.'' Before shows the bowls still
separated. The imagined trajectories illustrate different stacking orders and
verification outcomes; After shows the two successive stacking operations.}
\label{fig:case-bowls}
\end{figure}

\begin{figure}[!htbp]
\centering
\includegraphics[width=\linewidth,trim=0 132bp 0 132bp,clip]{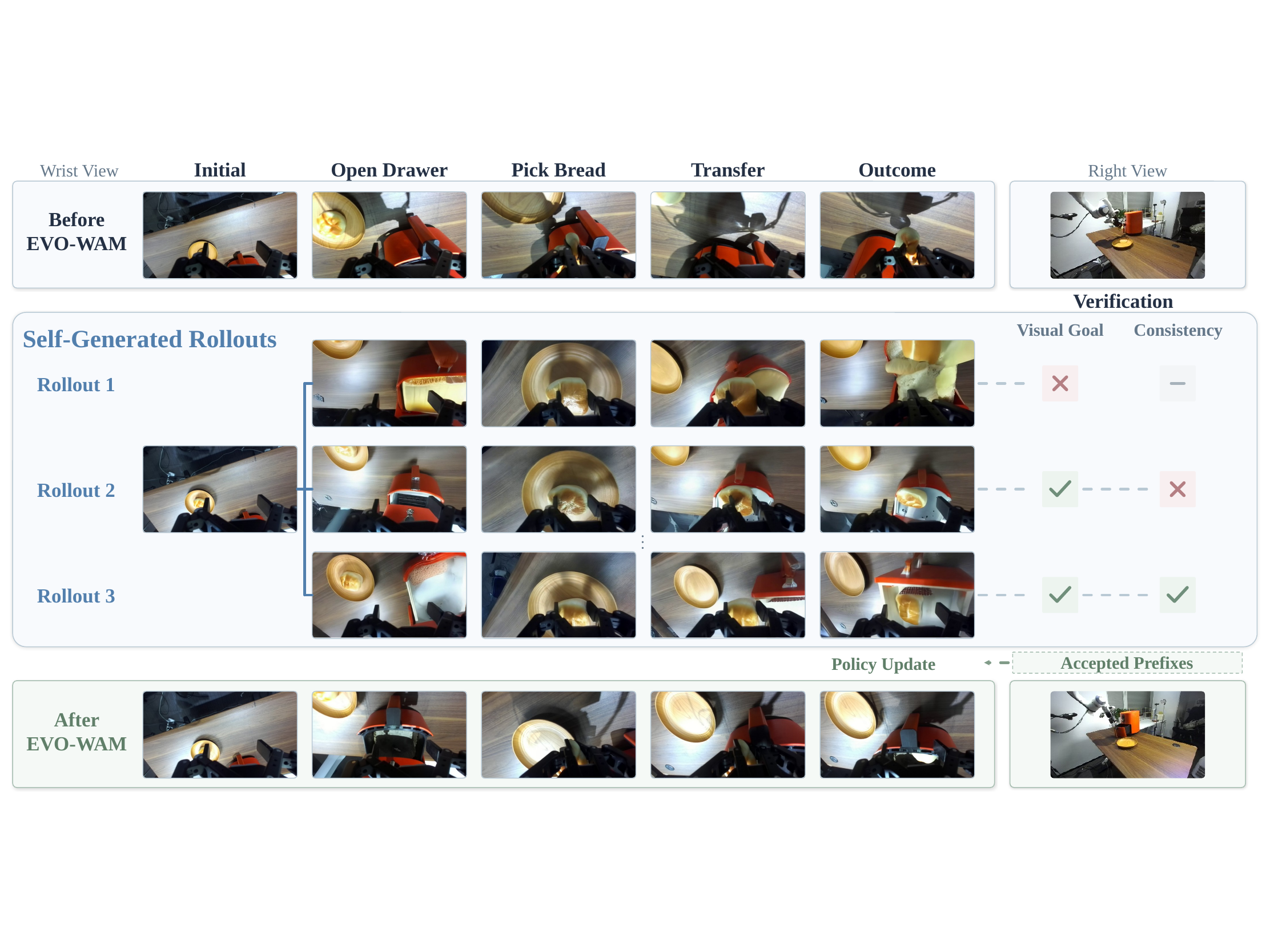}
\caption{\textbf{Loading the air fryer.} The robot must pull the red handle to
open the drawer and transfer the bread from the plate into it. Before shows the
bread still held by the gripper. Rollout 1 distorts the drawer and bread during
transfer, while Rollout 3 passes verification. After shows drawer opening,
bread transfer, and release.}
\label{fig:case-drawer}
\end{figure}

\paragraph{Qualitative case studies.}
Figures~\ref{fig:case-bowls} and~\ref{fig:case-drawer} show real executions before
self-training (Before), three self-imagined trajectories (Rollouts 1--3), and real
executions after the second self-training round (After). For each task, all three
self-imagined trajectories are generated from the same initial observation by the
model after its first self-training round. Rollout 1 is rejected by visual
verification. Rollout 2 passes the initial visual scan but is rejected by the IDM.
Rollout 3 passes both the IDM check and visual endpoint confirmation and is used
for the second round of self-training. Appendix~\ref{app:task-completion-details}
describes the verification procedure.

Each imagined trajectory is shown through four chronological wrist-camera frames.
The column headings indicate the intended task stages. The Before and After rows
show the initial observation and four subsequent wrist-camera frames, together
with a synchronized external view at the last displayed step. These real
executions use independently reset scenes with matching relative object layouts.

\paragraph{Stacking bowls: maintaining an intermediate result.}
Figure~\ref{fig:case-bowls} illustrates the dependency between the two stacking
operations. Rollout 1 places the white bowl above the pink and blue bowls,
reversing the required order. Rollout 2 places pink on blue and transfers the pair
onto white, ending with the gripper withdrawn. Its prefix IDM score of 0.008170
slightly exceeds the threshold of 0.008111, so the trajectory is rejected by the
action-consistency check. Rollout 3 passes verification and supplies a stacking
example for subsequent self-training rounds.

In Before, the bowls remain separated at the displayed endpoint. After shows the
robot placing pink onto blue and moving the resulting stack toward white. This
second operation requires preserving the intermediate stack while moving both
bowls together. The external view shows their relative positions alongside the
corresponding wrist view.

\paragraph{Loading the air fryer: opening before transfer.}
Figure~\ref{fig:case-drawer} illustrates how opening the drawer enables the
subsequent placement. In Before, the gripper still holds the bread at the
displayed endpoint. Rollout 1 develops a pronounced distortion in the drawer and
bread region after transfer. Rollout 2 opens the drawer, picks up the bread, and
releases it inside, leaving the plate empty. Its prefix IDM score is 0.009082,
above the threshold of 0.008111, so it is rejected. Rollout 3 passes verification
with a sequence that opens the drawer before transferring the bread.

In After, the robot first opens the drawer to make the receptacle accessible,
then grasps the bread, transfers it from the plate to the drawer, and releases it.
The sequence coordinates the prerequisite drawer interaction with object transfer.

\subsection{IDM Training Data and Model Design}
\label{app:training-ablations}

\paragraph{IDM design.}
The IDM predicts actions conditioned on observed motion, the window-start state,
and the instruction. Its training pairs include unsuccessful executions as well
as successful ones: in both cases, the recorded actions produced the observed
video. The IDM learns this correspondence by reconstructing the recorded actions. During verification, the generated video
conditions the reconstruction and the paired WAM actions are used to compute the
consistency error. Architecture, normalization, and temporal settings are given
in Appendix~\ref{app:idm-details}.

\subsection{Data Production and Training Configuration}
\label{app:configuration}

\paragraph{Baseline training.}
All simulation baselines are trained on the 43 seen RoboTwin tasks for
34K steps with a global batch size of 256. The learning rate follows
cosine decay over the first 15K steps, from $10^{-4}$ to $10^{-6}$
for Cosmos3 and the other comparison models, and from
$5\times10^{-5}$ to $5\times10^{-6}$ for DreamZero. At step 15K, the learning
rate is reset to $10^{-5}$ and kept constant for the remaining training steps.

\paragraph{Self-training optimization.}
Table~\ref{tab:self-training-config} lists the self-training settings. Each run
starts from a 30,000-step WAM checkpoint and initializes each round from the
preceding round's updated weights. RoboTwin uses four rounds of 1,000 updates;
the final checkpoints therefore have 34,000 training steps. The real-robot model
uses four rounds of 500 updates, reaching 32,000 steps; Table~\ref{tab:real-robot-main}
uses Round 2 at 31,000 steps. Batch size denotes
the global number of training samples per optimizer update.

\begin{table}[!htbp]
\centering\normalsize
\caption{\textbf{Self-training configuration.} Learning rates are constant within
each round. The recorded/generated ratio is the training sampling ratio.}
\label{tab:self-training-config}
\begin{tabularx}{\linewidth}{l>{\centering\arraybackslash}X>{\centering\arraybackslash}X>{\centering\arraybackslash}X}
\toprule
Setting & \methodbackbone{Cosmos3}, RoboTwin & \methodbackbone{DreamZero}, RoboTwin & \methodbackbone{Cosmos3}, real robot \\
\midrule
Starting step & 30,000 & 30,000 & 30,000 \\
Global batch size & 256 & 256 & 256 \\
Updates per round & 1,000 & 1,000 & 500 \\
Rounds & 4 & 4 & 4 \\
Learning rate & $10^{-5}$ & $2\times10^{-5}$ & $10^{-5}$ \\
Recorded/generated & 1:1 & 1:1 & 1:1 \\
Final training step & 34,000 & 34,000 & 32,000 \\
\bottomrule
\end{tabularx}
\end{table}

The action projection layers use a learning rate of $5\times10^{-5}$. In the
real-robot runs, generated prefixes,
recorded successful executions, and recorded unsuccessful executions contribute
50\%, 45\%, and 5\% of training samples, respectively. The generated-data sampler
selects a scene uniformly, an episode within that scene uniformly, and then a
chunk according to its duration. Training targets are the generated video and
its paired actions. IDM reconstructions provide the consistency score used for
filtering.

\paragraph{Verification-ablation controls.}
All variants in Table~\ref{tab:feedback} use the same Cosmos3 30K initialization,
a planned budget of 2,800 candidates per round, and four rounds of 1,000 updates.
They share the Cosmos3 learning rates and global batch size of 256 specified above,
the 1:1 recorded/generated sampling ratio, and cumulative prefix replay.
Each round is evaluated on the same seven tasks using 100 Clean and 100 Randomized
trials per task, following Section~\ref{sec:simulation-results}.
The action-verification comparison keeps the VLM, visual scanning, and endpoint
voting fixed: VLM only omits the action check, VLM + IDM uses consistency, and
VLM + Simulator uses task success from executing the paired actions.
The VLM comparison changes the task-completion model while retaining the same
IDM checkpoint and $\eta=0.00418487$. The retained prefix sets and their sizes
depend on the verifier's decisions; the training-update budget remains fixed.

\paragraph{Data retained per round.}
Table~\ref{tab:production-counts} reports newly retained prefixes, the
cumulative training pool, and optimizer updates. RoboTwin schedules 2,800
candidates per round from 1,400 distinct scenes across the seven unseen tasks.
In Round 2 of \methodbackbone{Cosmos3}, 2,799 candidates reached the recorded
selection stage. \methodbackbone{Cosmos3} and \methodbackbone{DreamZero} use fixed consistency thresholds of
0.00418487 and 0.00843549, respectively, throughout self-training, and accumulate
accepted prefixes. Once retained for training, prefixes remain in the pool in
subsequent rounds; the cumulative pool is the union of all rounds' retained sets.

\begin{table}[!htbp]
\centering\normalsize
\caption{\textbf{Retained training-pool sizes.} New prefixes are trajectories
retained for training after filtering and exclusions. Cumulative pool size sums
new prefixes through the current round. Each retained trajectory contributes one prefix. Updates are
optimizer steps per round.}
\label{tab:production-counts}
\begin{tabularx}{\linewidth}{>{\centering\arraybackslash}X>{\centering\arraybackslash}X>{\centering\arraybackslash}X>{\centering\arraybackslash}X}
\toprule
Round & New prefixes & Cumulative pool & Updates \\
\midrule
\multicolumn{4}{c}{\textit{\methodbackbone{Cosmos3} --- RoboTwin}} \\
1 & 531 & 531 & 1,000 \\
2 & 1,400 & 1,931 & 1,000 \\
3 & 1,482 & 3,413 & 1,000 \\
4 & 1,592 & 5,005 & 1,000 \\
\midrule
\multicolumn{4}{c}{\textit{\methodbackbone{DreamZero} --- RoboTwin}} \\
1 & 704 & 704 & 1,000 \\
2 & 894 & 1,598 & 1,000 \\
3 & 1,161 & 2,759 & 1,000 \\
4 & 1,234 & 3,993 & 1,000 \\
\midrule
\multicolumn{4}{c}{\textit{\methodbackbone{Cosmos3} --- Real robot}} \\
1 & 33 & 33 & 500 \\
2 & 159 & 192 & 500 \\
3 & 179 & 371 & 500 \\
4 & 76 & 447 & 500 \\
\bottomrule
\end{tabularx}
\end{table}

For the three real-robot tasks, each round starts with a budget of 100
candidates per scene, or 800 across eight scenes. Initially, a fallback procedure
adds batches of 100 for scenes with fewer than three newly retained prefixes.
This procedure is disabled during R2, with already submitted batches completed;
total generation attempts are 5,900 in R1 and 900 in R2.
R1 retains 33 training prefixes after excluding one corrupted generated video
from 34 automated acceptances. R2 adds 159 prefixes, yielding a cumulative pool
of 192. R3 and R4 each generate 800 candidates without additional batches,
retaining 179 and 76 new prefixes, respectively, for cumulative pools of 371 and 447.
The real-robot consistency threshold remains fixed at 0.00811050.

\subsection{Analysis of Generated Rollouts}
\label{app:rollout-analysis}

The candidates in Figures~\ref{fig:case-bowls} and~\ref{fig:case-drawer} share
an initial observation within each task, yet differ in stacking order and object
geometry. The following example examines a separate discrepancy: a generated
placement that does not occur when the paired actions are executed.

\begin{figure}[!htbp]
\centering
\includegraphics[width=\linewidth]{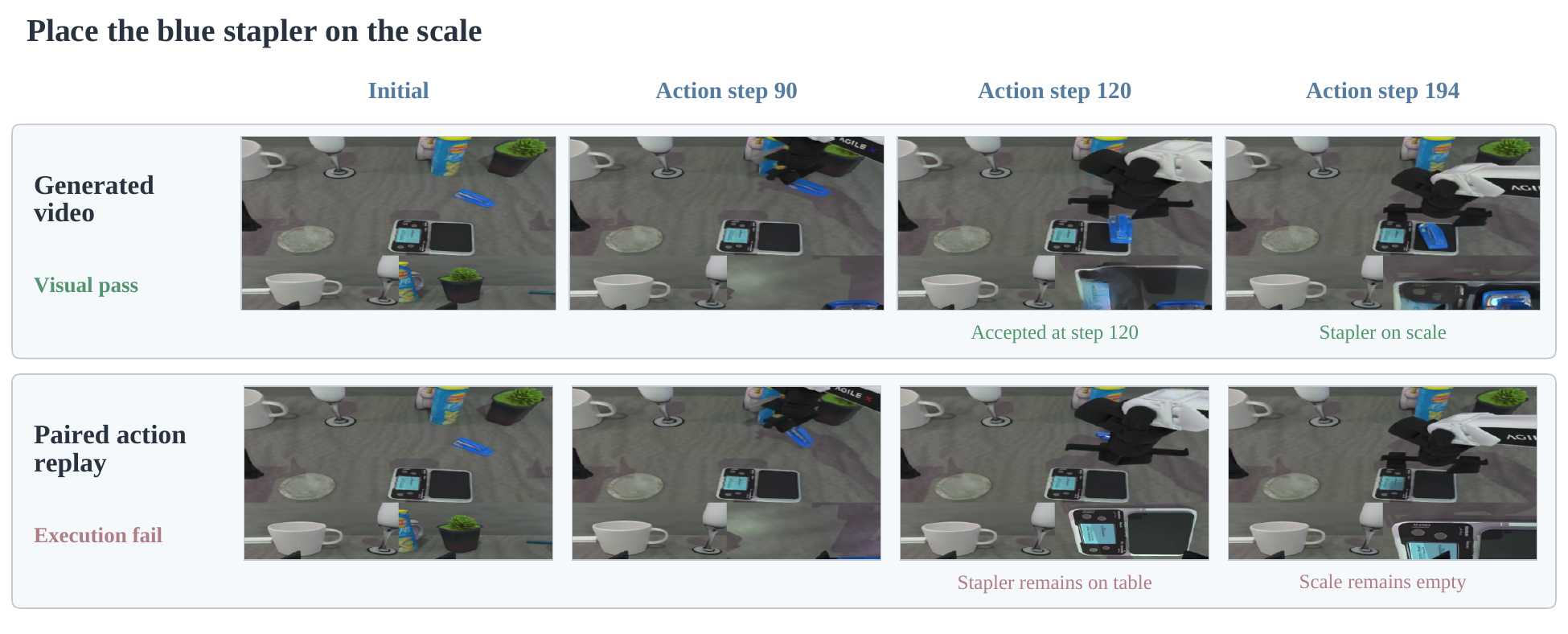}
\caption{\textbf{Generated completion and failed action execution.} Top: the
video generated by the 30,000-step Cosmos3 policy. Bottom: simulator execution
of the same generated action sequence. Columns use the same action steps and
show the main view above the two wrist views. The generated stapler reaches the
scale; the replayed stapler remains on the table. The last column is the end of
the 194-action replay. The visually accepted 120-action prefix has an IDM consistency
error of 0.00632283, exceeding the fixed threshold of 0.00418487, and is therefore
rejected by the IDM.}
\label{fig:paired-execution}
\end{figure}

Figure~\ref{fig:paired-execution} compares a generated rollout with execution of
its paired action sequence from the same initial scene. The task is to place the
blue stapler on the scale using the right arm. In the generated video, the stapler
moves onto the weighing platform. In the simulator replay, the gripper moves
toward the scale but leaves the stapler on the table. At action step 120, the
generated image shows the stapler on the scale while the replayed scale remains
empty. The stapler remains on the table through the end of the 194-action sequence,
and the simulator reports task failure.

This diagnostic rollout was generated by the starting policy before self-training.
Visual assessment accepts the generated placement at step 120, but executing the
paired WAM actions fails to grasp the stapler. The IDM rejects this prefix,
illustrating why visual completion needs to be checked together with
video--action consistency for each candidate prefix.

\subsection{Verifier Reliability}
\label{app:verifier-reliability}

We assess selection quality on 700 Cosmos3 rollouts sampled from a pool of
8,390 candidates with recorded simulator replay outcomes. We sample 50 per
task--condition pair across seven tasks and Clean/Randomized conditions, using
seed 20260926 independently of verifier decisions and replay labels. The subset
contains 208 successes and 492 failures. Both methods share the recorded action
prechecks, visual endpoints, and two-of-three VLM voting. VLM + IDM additionally
applies the IDM trained on the 43 seen tasks with fixed $\eta=0.00418487$.

\begin{table}[!htbp]
\centering\normalsize
\caption{\textbf{Verifier reliability against simulator replay (\%).}
Both methods use the same 700 candidates. Precision and recall treat replay
success as the positive label; FPR is the fraction of replay failures accepted.}
\label{tab:verifier-reliability}
\begin{tabular}{lccc}
\toprule
Verification & Precision $\uparrow$ & Recall $\uparrow$ & FPR $\downarrow$ \\
\midrule
VLM & 68.0 & 64.4 & 12.8 \\
\rowcolor{oursblue}
VLM + IDM & 86.0 & 44.2 & 3.0 \\
\bottomrule
\end{tabular}
\end{table}

As shown in Table~\ref{tab:verifier-reliability}, adding IDM reduces false
acceptances from 63 to 15, at the cost of lower recall.
Among VLM-accepted candidates, replay succeeds for 92/107 (86.0\%) passing IDM
and 42/90 (46.7\%) rejected by IDM, linking consistency verification to a higher
proportion of successful executions. These labels describe complete action-tape
replay, not separate execution of selected prefixes or human judgments of visual
completion. Simulator failures may also reflect physics artifacts and do not by
themselves identify video--action inconsistency; all recorded failures remain
in the analysis.

\section{Implementation Details}
\label{app:implementation-details}

\subsection{Task-Completion Verification}
\label{app:task-completion-details}

\paragraph{Visual assessment.}
The description call receives synchronized initial and generated views with
observation prompts, without the task instruction or desired outcome. It records
object identities, spatial relations, gripper contact, and visible changes in
object or robot structure. The judgment call then receives the same images,
the resulting description, the task instruction, and the active subgoal.
It checks goal satisfaction, required gripper release, object consistency,
and robot structural consistency. The program accepts an assessment only when
all four checks pass. A negative or uncertain check makes the assessment
non-accepting. Unsupported or conflicting facts in the description and its
image-grounded audit also prevent the affected check from passing.
Release is not required for the handle-engagement subgoal of opening the air fryer.

\paragraph{Sequential scanning.}
Both RoboTwin and real-robot verification use the same visual endpoint search.
For subgoals $(g_1,\ldots,g_M)$ checked in sequence, let $\mathcal T_j$ contain predefined
scan times within a task-specific window for subgoal $g_j$.
Only the active subgoal is assessed, including any earlier
relations it requires to remain satisfied. For placing ducks, the second subgoal
requires both the blue duck in the blue bowl and the pink duck still in the pink bowl.
Write $b_j^{(q)}(t)=1$ when assessment $q$ accepts and zero when a valid
assessment rejects or remains uncertain. The preliminary completion time is
\begin{equation}
t_j=\min\{t\in\mathcal T_j:t>t_{j-1},\ b_j^{(1)}(t)=1\},
\qquad t_0=0.
\label{eq:subgoal-endpoint}
\end{equation}
The scan advances to $g_{j+1}$ only after finding $t_j$. Each endpoint is selected
using the initial VLM assessment alone and remains fixed during the subsequent
IDM check and vote confirmation. For real-robot tasks,
scan windows are 7--17\,s and 17--33\,s for stacking bowls, 5--14\,s and
12--23\,s for placing ducks, and 1\,s to the rollout end and 19--36\,s for
loading the air fryer. The increasing-time constraint also applies where these
windows overlap. A rollout without a completion time for every subgoal is rejected.

\paragraph{Endpoint confirmation.}
After the prefix ending at $t_v=t_M$ passes the IDM check, we confirm each fixed
endpoint. The initial accepting assessment and two additional
description--judgment calls form three votes. Each additional vote receives the
same images and subgoal but produces its own description. A subgoal is confirmed
when
\begin{equation}
\sum_{q=1}^{3}b_j^{(q)}(t_j)\geq2,
\qquad j=1,\ldots,M.
\label{eq:subgoal-voting}
\end{equation}
A missing or invalid response is not counted as a negative vote.
Unresolved candidates are withheld from subsequent policy training.

\paragraph{Interaction with action verification.}
Algorithm~\ref{alg:prefix-verification} applies to both RoboTwin and real-robot tasks:
first locate and fix all visual endpoints, then check $E(t_M)\leq\eta$, and finally
confirm each endpoint with two additional VLM assessments. IDM or voting rejection
discards the candidate without searching for a later endpoint. Training retains
the complete prefix through $t_v$ only after both conditions pass; all its actions
are scored using the window rules in Appendix~\ref{app:idm-details}.

\begin{algorithm}[!htbp]
\caption{Prefix verification for RoboTwin and real-robot tasks}
\label{alg:prefix-verification}
\normalsize
\begin{algorithmic}[1]
\Require Rollout $\hat\tau$, ordered subgoals $(g_j)_{j=1}^M$, scan sets $(\mathcal T_j)_{j=1}^M$, fixed threshold $\eta$
\Statex Any unresolved assessment or unavailable IDM score returns \textsc{Withhold}.
\State $t_0\gets0$
\For{$j=1,\ldots,M$}
  \State $t_j\gets\bot$
  \For{$t\in\mathcal T_j$ in increasing order, with $t>t_{j-1}$}
    \State Obtain initial VLM assessment $b_j^{(1)}(t)$ from $\hat\tau$ for $g_j$
    \State \textbf{if} $b_j^{(1)}(t)=1$ \textbf{then} $t_j\gets t$; \textbf{break}
  \EndFor
  \State \textbf{if} $t_j=\bot$ \textbf{then return} \textsc{Reject}
\EndFor
\State Fix $t_v\gets t_M$ and all subgoal endpoints $(t_1,\ldots,t_M)$
\State Compute $E(t_v)$ over all actions through $t_v$ (Appendix~\ref{app:idm-details})
\State \textbf{if} $E(t_v)>\eta$ \textbf{then return} \textsc{Reject}
\For{$j=1,\ldots,M$}
  \State Obtain $b_j^{(2)}(t_j)$ and $b_j^{(3)}(t_j)$ using the same images and subgoal
  \State \textbf{if} $\sum_{q=1}^{3}b_j^{(q)}(t_j)<2$ \textbf{then return} \textsc{Reject}
\EndFor
\State \Return the complete prefix ending at $t_v$
\end{algorithmic}
\end{algorithm}

\begin{figure}[!htbp]
\centering
\includegraphics[width=\linewidth]{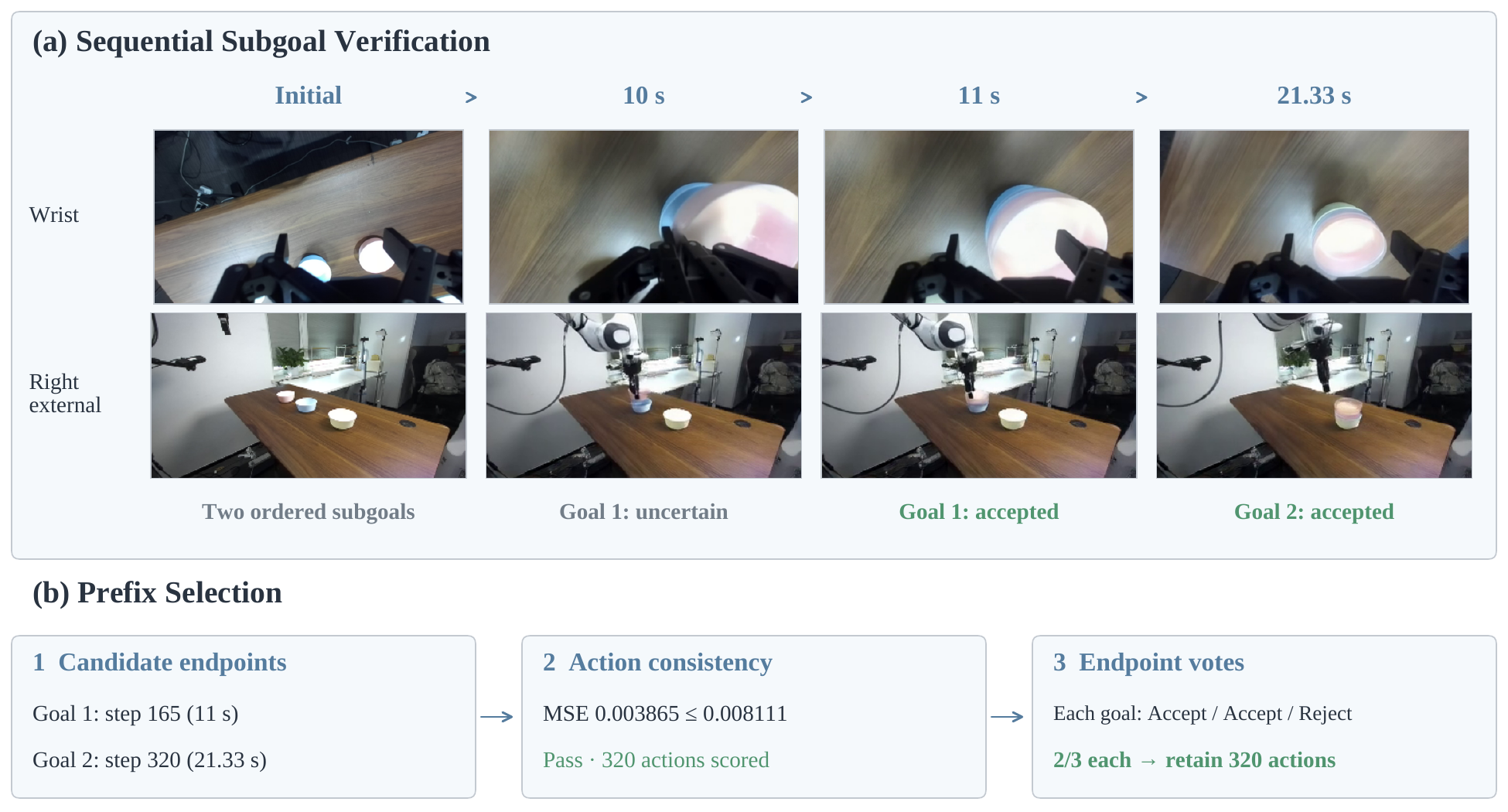}
\caption{\textbf{A recorded verification trace for stacking bowls.} The initial VLM
scan proposes endpoints at step 165 (11\,s) for the first subgoal and step 320
(21.33\,s) for the second.
The 320-action prefix passes the IDM
threshold, after which each endpoint receives two further assessments. Both
subgoals receive Accept/Accept/Reject, satisfying the two-of-three rule.
Five complete 64-action windows are scored, and all 320 actions are retained for training. The
images show wrist and right external views; assessment also uses the left view.
}
\label{fig:verification-trace}
\end{figure}

\subsection{Inverse Dynamics Model}
\label{app:idm-details}

\paragraph{Architecture and training data.}
The IDM reconstructs actions from a video window, its starting robot state, and the task
instruction. We build it from Wan2.2 pretrained weights~\citep{DBLP:journals/corr/abs-2503-20314} using a reduced-width Transformer
initialized by interpolating the pretrained tensors. We retain video conditioning and
action denoising, while removing the video denoising branch and its prediction loss. The
resulting model has approximately 0.9B trainable parameters. The Transformer has 30 layers,
a hidden width of 1,344, an FFN width of 5,376, and 24 attention heads. Video and
text encoders provide the conditioning features for action flow prediction.

We train the IDM on recorded video-action pairs, including successful and unsuccessful
executions: an unsuccessful task can still provide a valid correspondence between motion
and actions. For RoboTwin, gradient training uses recorded trajectories from the 43 seen tasks.
For real-world verification, we train the IDM on
DROID~\citep{DBLP:conf/rss/KhazatskyP0BDKN24} trajectories.

\paragraph{Temporal and training settings.}
Table~\ref{tab:idm-config} lists the three temporal interfaces. The dense RoboTwin
IDM is trained for 26,000 updates with global batch size 256. The sparse-video
IDM for DreamZero continues that model for 10,000 updates at batch size 256.
The DROID IDM uses a 50,000-update continuation of a pretrained DROID IDM, also
at batch size 256.

\begin{table}[!htbp]
\centering\normalsize
\caption{\textbf{IDM settings.} $L$ is the number of actions in a verification window;
the video includes its starting observation. Standardization uses recorded-data
means and standard deviations.}
\label{tab:idm-config}
\begin{tabularx}{\linewidth}{l>{\centering\arraybackslash}X>{\centering\arraybackslash}X>{\centering\arraybackslash}X}
\toprule
Setting & RoboTwin, Cosmos3 & RoboTwin, DreamZero & DROID, Cosmos3 \\
\midrule
Action dimensions & 14 & 14 & 8 \\
Actions per window & $1\leq L\leq64$ & $L\in\{3,6,\ldots,72\}$ & $1\leq L\leq64$ \\
Video frames & $L+1$ & $L/3+1$ & $L+1$ \\
Video/action rate & 15/15 Hz & 5/15 Hz & 15/15 Hz \\
Normalized clipping & None & None & $[-5,5]$ \\
Reconstruction steps & 4 & 4 & 4 \\
Calibration percentile & P90 & P90 & P99 \\
Fixed consistency threshold & 0.00418487 & 0.00843549 & 0.00811050 \\
\bottomrule
\end{tabularx}
\end{table}

The dense RoboTwin model samples 64-action windows with probability 0.5, 32-action windows
with probability 0.25, and other supported lengths with the remaining probability.
For generated rollouts, window-start conditioning uses the measured initial robot
state for the first window and the corresponding boundary-state estimate for
subsequent windows.

\paragraph{Training objective.}
Let $A$ be a training action sequence with length $L$ and dimension $d$, and let
$Y=\mathcal N_{\mathrm{IDM}}(A)$ be its normalized representation. Given Gaussian noise
$\epsilon$ and a sampled noise level $\sigma$, we construct
\begin{equation}
X_\sigma=(1-\sigma)Y+\sigma\epsilon.
\label{eq:idm-noising}
\end{equation}
The IDM is trained with an action flow-matching objective~\citep{DBLP:conf/iclr/LipmanCBNL23}
\begin{equation}
\mathcal L_{\mathrm{IDM}}
=
\mathbb E
\left[
\frac{1}{Ld}
\left\|
v_\phi(X_\sigma,\sigma;\widetilde Z,c,\ell)
-
(\epsilon-Y)
\right\|_F^2
\right],
\label{eq:idm-training}
\end{equation}
where $\phi$ denotes the IDM parameters, $c$ is the window-start state, and
$\widetilde Z$ is the encoded video condition. With probability 0.5, we perturb the video condition as
$\widetilde Z=(1-\alpha)Z+\alpha\epsilon_Z$, where
$\alpha\sim\mathcal U(0,0.5)$ and $\epsilon_Z$ is Gaussian noise. Otherwise,
$\widetilde Z=Z$.

\paragraph{Action reconstruction.}
Each reconstruction conditions on the corresponding video, including the window's starting
observation, its starting state, and the instruction. Reconstruction uses four UniPC steps~\citep{DBLP:conf/nips/ZhaoBR0L23}, starting from independent
Gaussian action noise; the WAM-generated actions are used only for the comparison.

\paragraph{Video-action consistency score.}
For a proposed prefix ending at $t_v$, let $\hat A_k^{(v)}$ contain the $n_k$ generated
actions in verification window $k$, and let $\widetilde Y_k^{(v)}$ denote the corresponding
IDM reconstruction in normalized action space. With $d$ denoting the action dimension and
$\mathcal N_{\mathrm{IDM}}$ the action normalization used by the IDM, the window error is
\begin{equation}
e_k
=
\frac{1}{n_kd}
\left\|
\mathcal N_{\mathrm{IDM}}\bigl(\hat A_k^{(v)}\bigr)
-
\widetilde Y_k^{(v)}
\right\|_F^2.
\label{eq:chunk-consistency}
\end{equation}
For both RoboTwin and DROID, the $K_v$ windows partition all $N_v$ actions through
$t_v$, including a shorter final window when needed. Thus $\sum_{k=1}^{K_v}n_k=N_v$,
and we aggregate window errors weighted by their respective numbers of actions:
\begin{equation}
E(t_v)
=
\frac{\sum_{k=1}^{K_v}n_ke_k}
     {\sum_{k=1}^{K_v}n_k}.
\label{eq:prefix-consistency}
\end{equation}
The complete prefix ending at $t_v$ is retained for training if task-completion verification
passes and $E(t_v)\leq\eta$, where $\eta$ is fixed across self-training rounds for each
backbone and dataset.

For Cosmos3 on RoboTwin and DROID, $K_v=\lceil N_v/64\rceil$ and the final window
contains $N_v-64(K_v-1)$ actions. This window is scored and weighted by its actual
action count, so the visual, scoring, and training endpoints all remain at $t_v$.
Candidate actions use the IDM's normalization,
including DROID clipping, and are compared directly with the normalized IDM output.

\paragraph{Threshold calibration.}
We set thresholds separately for simulation and real-robot verification to account
for differences in their video-action data distributions.
For both RoboTwin backbones, IDM checkpoint selection and threshold calibration
use offline data from the 43 seen tasks, without execution feedback from the
seven target tasks. We calibrate once per backbone, setting $\eta$ to the 90th
percentile (P90) of normalized video-action reconstruction errors on these
calibration data. The selected IDM and its numerical threshold remain fixed
across self-training rounds. For DROID, we use the 99th percentile (P99) of
normalized video-action reconstruction errors on the original recorded DROID
data, yielding $\eta=0.00811050$, which is also fixed across self-training rounds.
Table~\ref{tab:idm-config} reports the calibration percentiles and fixed thresholds
for all three settings.

\subsection{Autoregressive Rollout Training}
\label{app:rollout-training}

\paragraph{Chunk and context settings.}
Table~\ref{tab:autoregressive-config} separates generated outputs from reused
context. Cosmos3 predicts 64 video frames and 64 actions per chunk. DreamZero
predicts 24 video frames and 72 actions because video is sampled at 5\,Hz and
actions at 15\,Hz. At each chunk boundary, the state input for continuation is
obtained from the model's outputs and combined with the initial visual anchor
and recent generated latents.
The initial anchor is retained throughout the rollout, and generated latents
are reused directly.

\begin{table}[!htbp]
\centering\normalsize
\caption{\textbf{Autoregressive generation settings.} Output counts exclude the
starting observation. Recent visual context is in addition to the initial anchor.}
\label{tab:autoregressive-config}
\begin{tabularx}{\linewidth}{l>{\centering\arraybackslash}X>{\centering\arraybackslash}X>{\centering\arraybackslash}X}
\toprule
Setting & Cosmos3, RoboTwin & DreamZero, RoboTwin & Cosmos3, DROID \\
\midrule
Video/action rate & 15/15 Hz & 5/15 Hz & 15/15 Hz \\
New frames/actions & 64/64 & 24/72 & 64/64 \\
Recent video context & 32 frames & 8 frames & 32 frames \\
Recent video latents & 8 & 2 & 8 \\
\bottomrule
\end{tabularx}
\end{table}

For Cosmos3 training, a continuation sample uses 33 local frames: one VAE priming
frame and 32 recent frames. The priming latent is dropped before the eight history
latents are combined with the initial anchor. The stitched trajectory concatenates
the newly predicted outputs of successive chunks.

\paragraph{Training contexts.}
Both initialization and continuation modes are used during base training and each
self-training round. Training contexts and prediction targets are drawn from the same
trajectory: recorded trajectories during base training, and a mixture of recorded and
verified generated trajectories during self-training. During autoregressive generation,
the model directly reuses its predicted video latents and obtains the next
state input from its own outputs to construct the continuation context.

\Needspace{0.68\textheight}
\subsection{Qualitative Comparison of Visual Context}
\label{app:context-consistency}

\begin{figure}[H]
\centering
\includegraphics[width=\linewidth]{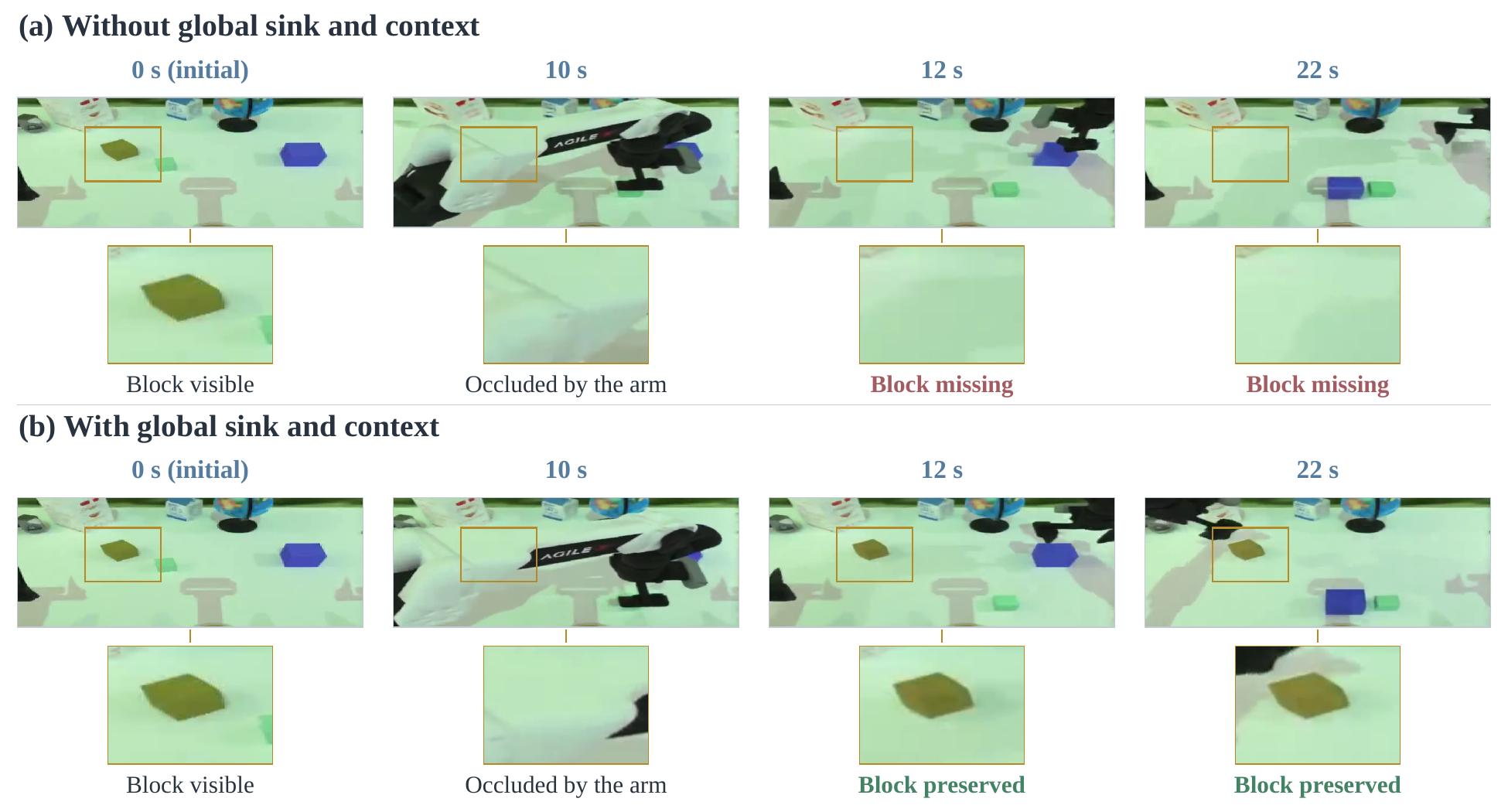}
\caption{\textbf{Object consistency during autoregressive rollout.}
Top: without global sink and context. Bottom: with both components.
Columns show matched timestamps relative to each clip's start. The yellow block
is initially visible and is occluded by the arm at 10\,s. At 12\,s and 22\,s,
it is missing in the top row and preserved in the bottom row.
Each frame shows the main camera view cropped from the source video; boxes and
enlarged insets highlight the same fixed image region in both rows.}
\label{fig:context-object-consistency}
\end{figure}

Figure~\ref{fig:context-object-consistency} compares two autoregressive rollouts
with and without the initial visual anchor (global sink) and recent multi-frame
context described in Section~\ref{sec:autoregressive-rollouts}. The comparison
illustrates object persistence after occlusion: the yellow block disappears in
the rollout without these context components, while it remains visible after
the arm moves away in the rollout with them.

\end{document}